\documentclass[journal]{IEEEtran}
\usepackage{cite}
\usepackage{amsmath,amssymb,amsfonts}
\usepackage{algorithmic}
\usepackage{graphicx}
\usepackage{multirow}
\usepackage{textcomp}
\usepackage{booktabs}
\usepackage[colorlinks=true]{hyperref}

\begin{document}
\title{GenSelfDiff-HIS: Generative Self-Supervision Using Diffusion for Histopathological Image Segmentation}
\author{Vishnuvardhan Purma$^\dag$, Suhas Srinath$^\dag$, Seshan Srirangarajan, Aanchal Kakkar, and Prathosh A.P.
\thanks{Vishnuvardhan Purma is with the Electrical Engineering Department, and Suhas Srinath and Prathosh A. P. are with the Electrical Communication Engineering Department, Indian Institute of Science (IISc), Bengaluru, Karnataka 560012 India (e-mails: \{purmav, suhass12, prathosh\}@iisc.ac.in). Seshan Srirangarajan is with the Electrical Engineering Department, Indian Institute of Technology Delhi (IIT-D), New Delhi 110016, India. Aanchal Kakkar is with the Department of Pathology, All India Institute of Medical Sciences Delhi (AIIMS), New Delhi 110029, India. $\dag$ Equal contribution.}}

\maketitle

\begin{abstract}
Histopathological image segmentation is a laborious and time-intensive task, often requiring analysis from experienced pathologists for accurate examinations. To reduce this burden, supervised machine-learning approaches have been adopted using large-scale annotated datasets for histopathological image analysis. However, in several scenarios, the availability of large-scale annotated data is a bottleneck while training such models. Self-supervised learning (SSL) is an alternative paradigm that provides some respite by constructing models utilizing only the unannotated data which is often abundant. The basic idea of SSL is to train a network to perform one or many pseudo or pretext tasks on unannotated data and use it subsequently as the basis for a variety of downstream tasks. It is seen that the success of SSL depends critically on the considered pretext task. While there have been many efforts in designing pretext tasks for classification problems, there have not been many attempts on SSL for histopathological image segmentation. Motivated by this, we propose an SSL approach for segmenting histopathological images via generative diffusion models. Our method is based on the observation that diffusion models effectively solve an image-to-image translation task akin to a segmentation task. Hence, we propose generative diffusion as the pretext task for histopathological image segmentation. We also utilize a multi-loss function-based fine-tuning for the downstream task. We validate our method using several metrics on two publicly available datasets along with a newly proposed head and neck (HN) cancer dataset containing Hematoxylin and Eosin (H\&E) stained images along with annotations. Code available at \url{https://github.com/suhas-srinath/GenSelfDiff-HIS}.

\end{abstract}

\begin{IEEEkeywords}
Diffusion, H\&E-stained Histopathological Images, Representation Learning, Self-Supervised Learning.
\end{IEEEkeywords}

\section{Introduction}
\label{sec:introduction}
\IEEEPARstart{A}utomated histopathological analysis has received a lot of attention owing to its utility in reducing time-intensive and laborious efforts of human pathologists \cite{b56, b57, b58}. 
Deep learning-based models are the ubiquitous choice for this purpose, which learn useful task-specific representations, enabling efficient mapping to the corresponding task labels \cite{b1}.
Existing methods \cite{b2, b59, b60, b78} for medical image segmentation focused on learning representations in a fully supervised manner, demanding substantial amounts of annotated data which is often difficult to obtain in histopathology. Moreover, biases and disagreements in annotations from experts lead to uncertainty in the ground truth labels themselves. This motivates the study of unsupervised learning, particularly self-supervised learning (SSL).
SSL uses the information available from a large number of unannotated images to learn effective visual representations through designing pseudo or pretext tasks \cite{b61, b62}.
These learned representations can then improve the performance of downstream tasks such as classification and segmentation with a limited number of labeled images, thereby reducing the amount of annotated data required for training the models.

The pretext tasks for SSL methods can be broadly classified as predictive, contrastive, and generative (refer Section \ref{rel_ssl}). Most of the existing SSL approaches \cite{b3, b4, b5, b6, b7, b8, b9, b10, b11, b12} are based on predictive and contrastive pretext tasks. Some methods like \cite{b13, b41} use a combination of predictive and contrastive tasks. However, generative pretext tasks of SSL have remained relatively unexplored, particularly in histopathology. Generative pretext tasks can potentially be more suitable for histopathological segmentation tasks since they are tasked to model the entire image distribution which is conducive to a downstream segmentation task. On the contrary, in predictive and contrastive SSL, the designed pretext task focuses on learning `salient features' without necessarily learning the entire image distribution needed for segmentation tasks. This motivates us to explore the use of generative models for SSL of histopathological images.  

Previous attempts on generative SSL \cite{b14, b15, b16, b17} utilize models such as variational autoencoders (VAEs) \cite{b18} and generative adversarial networks (GANs) \cite{b19}. However, GANs and VAEs suffer from several issues such as training instability, degraded image quality, and mode collapse \cite{b63}.
Recently, denoising diffusion probabilistic models (DDPMs) \cite{b44} have emerged to be powerful alternatives to GANs and VAEs in producing high-quality images \cite{b20}, which motivates us to use them as a pretext task. 
Additionally, since DDPMs inherently solve an image-to-image translation task using a segmentation-like backbone, they make a natural choice for self-supervised pretraining of segmentation problems. 
While DDPMs have been explored for medical image segmentation \cite{b21, b22} for modalities such as magnetic resonance imaging (MRI), computed tomography (CT), and ultrasound imaging modalities, the use of DDPMs specifically for SSL in the context of histopathological images is unexplored hitherto, to the best of our knowledge. 
While diffusion models have been used for SSL, they have not been used as generators in the context of segmentation, which we leverage in our work. 
Motivated by the aforementioned observations, we propose the use of DDPMs as pretext tasks in SSL for histopathological segmentation. Specifically, the contributions of our work can be summarized as follows:
\begin{enumerate}
    \item We propose to employ a UNet-based, image-space generative diffusion process as the pretext task in self-supervised learning for histopathological image segmentation.
    \item We propose to utilize the same UNet obtained from the aforementioned diffusion model, trained in a fully unsupervised manner, for segmentation. Specifically, the pre-trained diffusion UNet (step 1) is fine-tuned with segmentation losses with supervision for the final segmentation task.
    \item We propose a new head and neck (HN) cancer dataset with Hematoxylin and Eosin (H\&E) stained histopathological images along with corresponding segmentation mask annotations. To aid self-supervised and generative methods, the dataset contains a large number of unannotated images.
    \item We show the efficacy of our method on three histopathology datasets over multiple evaluation metrics with improved performance over other SSL pretext tasks and end-to-end supervised methods. Our framework also outperforms fully supervised DDPM-based methods and does not require large inference times. 
\end{enumerate}

\section{Related Work}

\subsection{Histopathological Image Analysis}
Deep convolutional networks have shown remarkable performance in histopathology, particularly in the segmentation of H\&E stained histopathological images. 
A comprehensive survey paper \cite{b23} delves into the methodological aspect of various machine learning strategies, including supervised, weakly-supervised, unsupervised, transfer learning, and their sub-variants within the context of histopathological image analysis. 
Komura \emph{et al.} \cite{b24} discuss diverse machine-learning methods for histopathological image analysis. 
Successful approaches such as U-Net-based networks \cite{b2, b25} have employed skip-connections between encoder and decoder parts to address the vanishing gradient problem similar to \cite{b26}.
These skip connections facilitate the extraction of richer feature representations. 
Xu \emph{et al.} \cite{b27} propose a novel weakly-supervised learning method called multiple clustered instance learning (MCIL) for histopathological image segmentation. 
MCIL performs image-level classification, medical image segmentation, and patch-level clustering simultaneously.
Another significant contribution is the fully convolutional network (FCN) based method \cite{b28}, which introduced a deep contour-aware network specifically designed for histopathological gland segmentation.
This method effectively tackles multiple critical issues in gland segmentation. 

Liu \emph{et al.} \cite{b29} presented a unique weakly-supervised segmentation framework based on sparse patch annotation for tumor segmentation. 
Bokhorst \emph{et al.} \cite{b30} compared two approaches, namely instance-based balancing and mini-batch-based balancing, when dealing with sparse annotations. 
Their study demonstrated that employing a large number of sparse annotations along with a small fraction of dense annotations yields performance comparable to full supervision. 
Yan \emph{et al.} \cite{b31} proposed a multi-scale encoder network to extract pathology-specific features, enhancing the discriminative ability of the network. 
Yang \emph{et al.} \cite{b32} present a deep metric learning-based histopathological image retrieval method that incorporates a mixed attention mechanism and derives a semantically meaningful similarity metric. 
An introductory and detailed review of histopathology image analysis is available in \cite{b33}. 
A recent study \cite{b34} provided a comprehensive survey of weakly-supervised, semi-supervised, and self-supervised techniques in histopathological image analysis.

\subsection{Self-Supervised Learning}\label{rel_ssl}
Self-supervised learning (SSL) is a paradigm aimed at learning visual feature representations from a large amount of unannotated data. The idea is to train a network to solve one of many pretext tasks (also known as pre-training) using the unlabelled data followed by a fine-tuning stage, where the model is further trained using a limited amount of annotated data for specific downstream tasks, such as nuclei and HN cancer histopathological image segmentation. This approach leverages the power of unsupervised learning to capture meaningful representations from unlabelled data and enhances the performance of supervised models in limited annotated data scenarios. 

The success of self-supervised models relies heavily on the choice of pretext tasks during the pre-training stage. 
Examples of pretext tasks include cross-channel prediction \cite{b3}, image context restoration \cite{b4}, image rotation prediction \cite{b5}, image colorization \cite{b6}, image super-resolution \cite{b7}, image inpainting \cite{b8}, and resolution sequence prediction (RSP) \cite{b9}. The quality of the learned visual features depends on the objective function of the pretext tasks and the pseudo labels generated from the available unannotated data. These pseudo-labels act as supervisory signals during the pre-training phase. Pretext tasks can be categorized into predictive tasks \cite{b9}, generative tasks \cite{b14, b15, b16}, contrasting tasks \cite{b10, b11, b12}, or a combination of them \cite{b13}. Predictive SSL is based on predictive tasks that focus on predicting certain properties or transformations of the input data. Generative SSL is based on generative tasks that involve generating plausible outputs from corrupted or incomplete inputs. Finally, contrastive SSL is based on contrasting tasks that aim to learn invariant representations under different augmentations of the same image. 

Jing \emph{et al.} \cite{b35} provide a detailed review of deep learning-based self-supervised general visual feature learning methods from images. Liu \emph{et al.} \cite{b36} reviewed comprehensively the existing empirical methods of self-supervised learning. 
Koohbanani \emph{et al.} \cite{b37} introduced novel pathology-specific self-supervision tasks that leverage contextual, multi-resolution, and semantic features in histopathological images for semi-supervised learning and domain adaptation. 
However, their study is limited to classification tasks. 
 
Contrastive learning methods aim to discriminate between instances and learn effective representations that capture essential characteristics \cite{b39, b54, b55}.
These methods extract the information in unannotated images by treating each unannotated image as a positive pair with its counterpart supervisory signal obtained through some transformation and considering the supervisory signals of other unannotated images as negative pairs.
Chaitanya \emph{et al.} \cite{b38} addressed the challenge of limited annotations in medical image segmentation through contrastive learning of global and local features on three MRI datasets.
Chen \emph{et al.} \cite{b39} demonstrated that incorporating a learnable non-linear transformation between the representations and contrastive loss can enhance the representation quality. 
However, they limited their study to natural images. 
In histopathology, Ciga \emph{et al.} \cite{b40} applied self-supervised contrastive learning to large-scale studies involving $57$ histopathological datasets. 
They observed improved performance across multiple downstream tasks like classification, regression, and segmentation through unannotated images.

A recent study \cite{b41} used the approach of cross-stain prediction and contrastive learning (CS-CO), which integrates the advantages of both predictive and contrastive SSL.
Xu \emph{et al.} \cite{b11} proposed a self-supervised deformation representation learning (DRL) approach, which uses elastically deformed images as the supervisory signals in the pre-training. This approach is based on maximizing the mutual information between the input images and the generated representations. In a recent study, Stacke \emph{et al.} \cite{b42} demonstrated the potential of contrastive self-supervised learning for histopathology applications in learning effective visual representations. Our method differs from the existing literature in that it leverages the potential of unannotated images through a generative self-supervision using diffusion as the pretext task.

\subsection{Diffusion Probabilistic Model for Image Segmentation}\label{rel_ddpm}
A diffusion probabilistic model consists of a fixed forward process where the input data undergoes perturbation to produce noisy data through a series of steps, and a learnable reverse process where the noise is progressively removed to recover the original noise-free input data. In recent years, diffusion models have gained significant attention in medical image analysis due to their remarkable success in various computer vision tasks. Diffusion models have been explored in tasks like image generation, image-to-image translation, reconstruction, classification, segmentation, denoising, etc.

Amit \emph{et al.} \cite{b71} were the first to apply diffusion models to image segmentation. 
They proposed to learn a conditional diffusion model using the input image. 
Baranchuk \emph{et al.} \cite{b74} showed that the representations learned by diffusion models effectively capture high-level semantic information which is used for label-efficient semantic segmentation. They perform segmentation as a pixel-wise classification by training ensembles of multi-layer perceptions (MLPs) for each pixel. 
Wolleb \emph{et al.} \cite{b73} proposed a semantic segmentation method where they employed a stochastic sampling process to generate a distribution of segmentation masks. 
Kim \emph{et al.} \cite{b75} introduced a novel diffusion adversarial representation learning model that leverages the diffusion model along with adversarial learning for vessel segmentation. 
In \cite{b72}, a framework known as collectively intelligent medical diffusion (CIMD) was introduced, that effectively captures the heterogeneity of segmentation masks without relying on an additional network for prior information during inference.
They also introduced a new metric for assessing both the accuracy and diversity of segmentation predictions, aligning with the interest of clinical practice of collective insights. 
Wu \emph{et al.} \cite{b21} proposed a diffusion-based model, called MedSegDiff, for general medical image segmentation.
They introduced a dynamic conditional encoding strategy for step-wise attention and proposed to eliminate high-frequency noise components. 
As an upgrade, Wu \emph{et al.} \cite{b22} proposed a novel Transformer-based Diffusion framework, called MedSegDiffV2. 
They employed different conditioning techniques over the backbone using raw images in the diffusion process.

Croitoru \emph{et al.} \cite{b76} provided a comprehensive review of articles on denoising diffusion models applied to vision, comprising both theoretical and practical contributions in the field. 
Kazerouni \emph{et al.} \cite{b77} provided a comprehensive overview of diffusion models within the field of medical imaging. 
Our framework is different from the aforementioned diffusion-based methods which are either explored for natural images or medical imaging modalities like CT, MRI, and ultrasound. 
Moreover, existing methods rely on the sampling process of a DDPM (consequently requiring an ensemble of predictions), introducing uncertainty (higher output variance) into the predicted segmentation masks.

\section{Methodology}
\subsection{Problem Formulation}
Let $\mathcal{S}_{\text{pr}}$ denote the set of unlabelled images used for SSL pre-training. Subsequently, a small set of labeled images, denoted by $\mathcal{S}_{\text{tr}}$ are used in the supervised segmentation task. 
Specifically, $\mathcal{S}_{\text{tr}}$ consists of both labeled images $\mathbf{x} \in \mathcal{X}$ and the corresponding masks $\mathbf{y} \in \mathcal{Y}$, where $\mathcal{X}$ and $\mathcal{Y}$ respectively denote the image and label spaces. 
Finally, the performance of the model is evaluated on a test set $\mathcal{S}_{\text{te}}$.
In our formulation, we assume $\mathcal{S}_{\text{pr}} \cap \mathcal{S}_{\text{tr}} = \phi$ (i.e., the unlabeled and labeled images are mutually exclusive). 
However, we also conduct experiments where $\mathcal{S}_{\text{pr}} = \mathcal{S}_{\text{tr}}$.
In the SSL stage, the task is to learn a representation function $f_{\theta}:\mathcal{S}_{\text{pr}}\to\mathcal{Z}$ (i.e., $f_{\theta}(\mathbf{x})=\mathbf{z}$), from the image space to latent space $\mathcal{Z}$, that would be effective during the downstream tasks. In this work, we propose to learn $f_{\theta}$ via a generative diffusion process described next. 

\subsection{Self-Supervision using Diffusion}
Let $\mathbf{x}_0\in \mathcal{S}_{\text{pr}}$ be the original input image and $\mathbf{x}_t$ be the corresponding noisy image obtained at time step $t=1,2, \dots, T$. 
Each $\mathbf{x}_t$ is obtained via $\mathbf{x}_{t-1}$ according to the following diffusion process:
\begin{equation}\label{eq2}
    \mathbf{x}_t = \sqrt{1-\beta_t}\mathbf{x}_{t-1} + \sqrt{\beta_t}\epsilon_t,
\end{equation}
where $\beta_t$ is the noise schedule parameter at time $t$, and $\epsilon_t \sim \mathcal{N}(\mathbf{0},\mathbf{I})$,  $\forall t=1,2, \dots, T$. 
This model imposes a set of encoding distributions $q(\mathbf{x}_t|\mathbf{x}_{t-1})$, which are assumed to be following a first-order Gaussian Markov process. 
In DDPMs, the encoding or the forward process is assumed to be fixed (i.e., $q(\mathbf{x}_{t}|\mathbf{x}_{t-1}) \sim \mathcal{N}(\mathbf{x}_{t}; \sqrt{1-\beta_t}\mathbf{x}_{t-1},\beta_t \mathbf{I})$), while the reverse or the decoding process is modeled using a parametric family of distributions denoted by $p_\theta(\mathbf{x}_{t-1}|\mathbf{x}_{t})$. 

The recursive formulation of $\mathbf{x}_t$ in \eqref{eq2} allows to sample the $\mathbf{x}_t$ directly from the original input image $\mathbf{x}_0$ as shown below:
\begin{equation}\label{eq3}
    \mathbf{x}_t = \sqrt{\Bar{\alpha}_t}\mathbf{x}_{0} + \sqrt{1-\Bar{\alpha}_t}\epsilon,
\end{equation}
where $\Bar{\alpha}_t=\prod_{i=1}^{t} \alpha_i$ and $\alpha_t=1-\beta_t$. If the noise schedule parameters $(\beta_t)_{t=1}^{T}$ are very small such that $\beta_T \rightarrow 0$, then the distribution of $\mathbf{x}_T$ can be well approximated by the standard Gaussian distribution i.e., $q(\mathbf{x}_T) \sim \mathcal{N}(\mathbf{0},\mathbf{I})$. 
The objective of the DDPM is to estimate the parameters of $p_\theta$, which is accomplished by optimizing a variational lower bound on the log-likelihood of the data $\mathbf{x}_0$ under the model $p_\theta$.

\begin{figure}[!t]
\centerline{\includegraphics[width=\columnwidth]{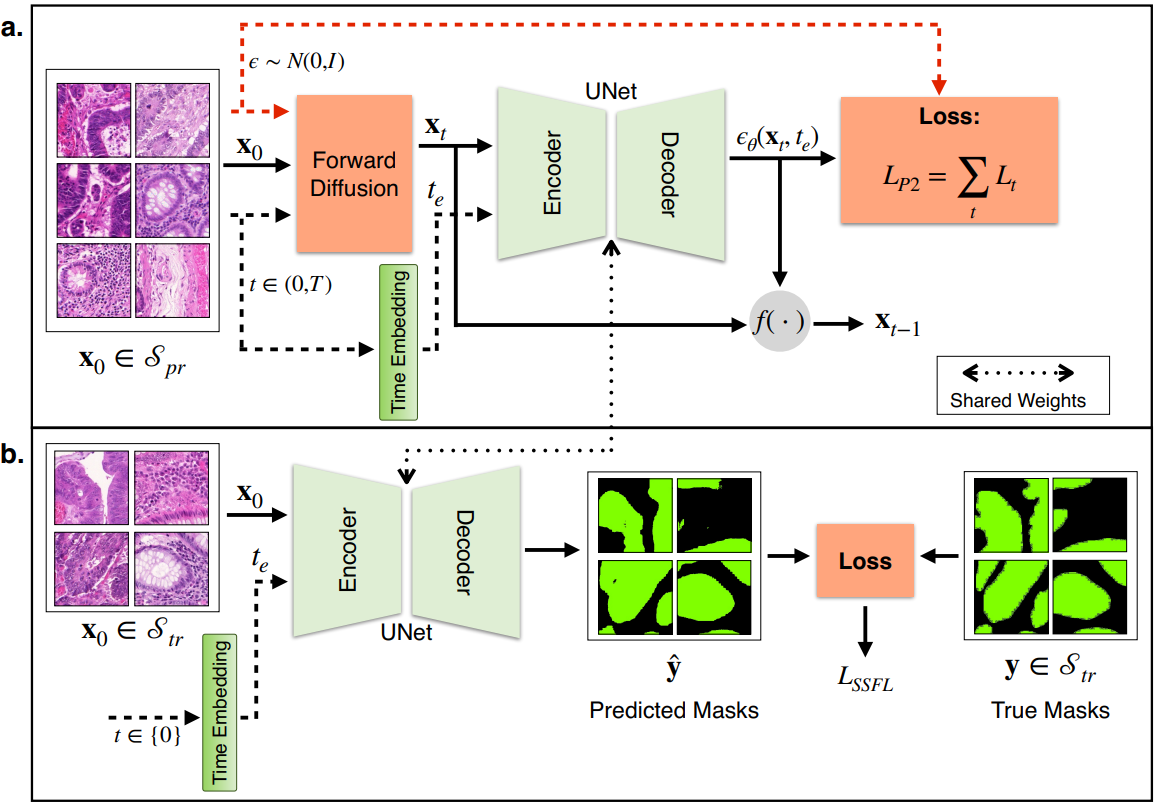}}
\caption{An overview of the proposed framework. (a) Self-supervised pre-training using diffusion: The U-Net model (encoder-decoder) takes the corrupted version $\mathbf{x}_t$ of the image $\mathbf{x}_0$ and the corresponding time embedding $t_e$ as the input to predict the noise that takes $\mathbf{x}_0$ to $\mathbf{x}_t$, using the P2 weighted \cite{b45} loss. $f(\cdot)$ denotes the function that recovers $\mathbf{x}_{t-1}$ from $\mathbf{x}_t$. (b) Downstream segmentation: The self-supervised pre-trained U-Net is fine-tuned end-to-end in a supervised manner to predict the segmentation masks.}
\label{fig:framework}
\end{figure}

Let $\mathbb{D}_{KL}[\cdot||\cdot]$ denote the Kullback-Leibler (KL) divergence between probability density functions.
The variational lower bound is shown to take the following form in \cite{b44,b43}:
\begin{align}\label{eq4}
    \log p_{\boldsymbol{\theta}}(\mathbf{x}_{0}) & \geq \underset{q(\mathbf{x}_{1}|\mathbf{x}_{0})}{\mathbb{E}}\bigr[\log p_{\boldsymbol{\theta}}(\mathbf{x}_{0}|\mathbf{x}_{1}) \bigr] - \mathbb{D}_{KL}\bigr[q(\mathbf{x}_{T}|\mathbf{x}_{0})||p(\mathbf{x}_{T})\bigr] \nonumber \\
    & - \sum_{t=2}^{T}\mathbb{D}_{KL}\bigr[q(\mathbf{x}_{t-1}|\mathbf{x}_{t}, \mathbf{x}_{0})||p_{\boldsymbol{\theta}}(\mathbf{x}_{t-1}|\mathbf{x}_{t})\bigr],
\end{align} 
where $q(\mathbf{x}_{t-1}|\mathbf{x}_{t}, \mathbf{x}_{0})  \sim \mathcal{N}\Bigl(\mathbf{x}_{t-1}; \mu_{q}(\mathbf{x}_{t}, \mathbf{x}_{0}), \sigma_{q}^{2}(t)\mathbf{I} \Bigl)$, $p_{\boldsymbol{\theta}}(\mathbf{x}_{t-1}|\mathbf{x}_{t}) \sim \mathcal{N}\Bigl(\mathbf{x}_{t-1}; \mu_{\boldsymbol{\theta}}, \sigma_{q}^{2}(t)\mathbf{I}\Bigl)$, 
\begin{align}
    \mu_{q}(\mathbf{x}_{t}, \mathbf{x}_{0}) & = \frac{\sqrt{\alpha_{t}}(1-\Bar{\alpha}_{t-1})\mathbf{x}_{t} + \sqrt{\Bar{\alpha}_{t-1}}(1-\alpha_{t})\mathbf{x}_{0}}{1-\Bar{\alpha}_{t}}, \nonumber \\
    \text{and } \sigma_{q}^{2}(t) & = \frac{(1-\alpha_{t})(1-\Bar{\alpha}_{t-1})}{1-\Bar{\alpha}_{t}}. \nonumber
\end{align}
$\mu_{\boldsymbol{\theta}}$ is the set of learnable model parameters. Equation \eqref{eq4} can be simplified by using the distributional forms (Gaussian) of $q(\mathbf{x}_{t-1}|\mathbf{x}_{t}, \mathbf{x}_{0})$ and $p_{\boldsymbol{\theta}}(\mathbf{x}_{t-1}|\mathbf{x}_{t})$ in $\mathbb{D}_{KL}\bigr[q(\mathbf{x}_{t-1}|\mathbf{x}_{t}, \mathbf{x}_{0})||p_{\boldsymbol{\theta}}(\mathbf{x}_{t-1}|\mathbf{x}_{t})\bigr]$, as follows:
\begin{align}\label{eq5}
    \log p_{\boldsymbol{\theta}}(\mathbf{x}_{0}) & \geq \underset{q(\mathbf{x}_{1}|\mathbf{x}_{0})}{\mathbb{E}}\bigr[\log p_{\boldsymbol{\theta}}(\mathbf{x}_{0}|\mathbf{x}_{1}) \bigr] \nonumber \\
    & - \mathbb{D}_{KL}\bigr[q(\mathbf{x}_{T}|\mathbf{x}_{0})||p(\mathbf{x}_{T})\bigr]  - \sum_{t=2}^{T}\frac{\| \mu_{q}(\mathbf{x}_{t}, \mathbf{x}_{0}) - \mu_{\boldsymbol{\theta}} \|^{2}}{2\sigma_{q}^{2}(t)}
\end{align}

While the above formulation is sufficient for model learning, it has been found that alternative noise-based reparameterization yields better performance \cite{b44}. Specifically, the third term in \eqref{eq5} can be re-parameterized \cite{b44} in terms of the `real' ($\epsilon$) and predicted ($\epsilon_\mathbf{\theta}$) noise parameters as follows:
\begin{align}\label{eq6}
    \log p_{\boldsymbol{\theta}}(\mathbf{x}_{0}) & \geq \underset{q(\mathbf{x}_{1}|\mathbf{x}_{0})}{\mathbb{E}}\bigr[\log p_{\boldsymbol{\theta}}(\mathbf{x}_{0}|\mathbf{x}_{1}) \bigr] - \mathbb{D}_{KL}\bigr[q(\mathbf{x}_{T}|\mathbf{x}_{0})||p(\mathbf{x}_{T})\bigr] \nonumber \\
    & - \sum_{t=2}^{T}\frac{(1-\alpha_{t})}{2\alpha_{t}(1-\Bar{\alpha}_{t-1})}\| \epsilon - \epsilon_{\boldsymbol{\theta}}(\mathbf{x}_{t}, t) \|^{2}.
\end{align}

A DDPM is optimized using the aforementioned formulation. While this formulation is shown to work on natural images, we empirically observe that this reparameterization improves the overall segmentation performance even in the case of histopathological images. Specifically, the first term in \eqref{eq6}, is a reconstruction term similar to the one obtained in vanilla VAE \cite{b18}. The second term is the prior matching term which represents the closeness of the noisy distribution $p(\mathbf{x}_{T})$ with the standard normal distribution. However, it is independent of network parameters $\boldsymbol{\theta}$ and hence can be neglected in the optimization. The third term is the denoising matching term which represents the closeness of desired denoising transition step $p_{\boldsymbol{\theta}}(\mathbf{x}_{t-1}|\mathbf{x}_{t})$ with the ground truth denoising transition step $q(\mathbf{x}_{t-1}|\mathbf{x}_{t}, \mathbf{x}_{0})$. 
Therefore, the resultant loss for training the DDPM is given by:
\begin{align}\label{eq7}
    L_{\text{vlb}} &= {\mathbb{E}} \bigr[ \text{log } p_{\boldsymbol{\theta}}(\mathbf{x}_{0}) \bigr] \approx \sum_{t \geq 1} L_{t} \nonumber \\
    L_{t} &= \underset{\mathbf{x}_{0}, \epsilon}{{\mathbb{E}}} \Bigr[\frac{(1-\alpha_{t})}{2\alpha_{t}(1-\Bar{\alpha}_{t-1})}\| \epsilon - \epsilon_{\boldsymbol{\theta}}(\mathbf{x}_{t}, t) \|^{2}\Bigr].
\end{align}

In practice, the weight factor in \eqref{eq7} is discarded to obtain the following simplified loss function:
\begin{align}\label{eq8}
    L_{\text{simple}} &= \sum_{t \geq 1} L_{t} \nonumber \\
    L_{t} &= \underset{\mathbf{x}_{0}, \epsilon}{{\mathbb{E}}} \Bigr[\| \epsilon - \epsilon_{\boldsymbol{\theta}}(\mathbf{x}_{t}, t) \|^{2}\Bigr]
\end{align}
Since the time-dependent weighting is discarded, the objective in \eqref{eq8} focuses on more difficult denoising tasks at larger $t$.

Our focus is mainly on learning effective visual representations using diffusion in a self-supervised manner. 
This can be achieved by focusing more on the content and avoiding insignificant or trivial details of the image. 
But, the simple diffusion loss or variational bound in \eqref{eq8} does not guarantee much because of uniform weighting for all the time steps. 
Choi \emph{et al.} \cite{b45} proposed a perception prioritized (P2) weighting and demonstrated that P2 weighting provides a good inductive bias for learning rich visual concepts by boosting weights at the coarse and the content stage and suppressing the weights at the clean-up stage (say high SNR or initial time steps). 
The variational bound with the P2 weighting scheme is
\begin{align}\label{eq9}
    L_{\text{P2}} &= \sum_{t \geq 1} L_{t}, \nonumber \\
    L_{t} &= \underset{\mathbf{x}_{0}, \epsilon}{{\mathbb{E}}} \Bigr[\frac{1}{(k+\text{SNR}(t))^{\gamma}} \| \epsilon - \epsilon_{\boldsymbol{\theta}}(\mathbf{x}_{t}, t) \|^{2}\Bigr],
\end{align}
where $\gamma$ is a hyperparameter that controls the strength of down-weighting focus on learning imperceptible details (high SNR). Here, $k$ is also a hyperparameter that prevents exploding weights for extremely small SNRs and determines the sharpness of the weighting scheme. SNR of the noisy sample $\mathbf{x}_{t}$ is obtained by taking the ratio of the squares of coefficients of $\mathbf{x}_{0}$ and $\epsilon$, corresponding to signal and noise variances, respectively, i.e., $\text{SNR}(t) = \frac{\Bar{\alpha}_t}{1-\Bar{\alpha}_t}$.

We finally use the P2 weighting loss \eqref{eq9} for the DDPM training (hyperparameter details are described later). It is to be noted that, architecturally, DDPM solves a regression task using a U-Net-like base network, which takes the noisy image (along with the time embedding) as the input to predict the noise content as shown in Fig. \ref{fig:framework}(a).
We use all the available unannotated images from the set $\mathcal{S}_{\text{pr}}$ to train the network in a self-supervised manner to obtain a rich set of representations $f_{\theta}$ for the input data distribution. Note that our objective of training a DDPM is not data generation but self-supervised representation learning. 

Post-training, we propose to use the same base U-Net at timestamp $t=0$ (obtained via DDPM training) to fine-tune for the downstream H\&E stained histopathological image segmentation tasks (refer Fig. \ref{fig:framework}(b)). 
The timestamp is added as an embedding layer with $t \in \{0\}$ since the downstream task does not involve or require noisy predictions. In other words, $\mathbf{x}_{0}$ is enough without any need for the noisy versions $\mathbf{x}_{1}, \mathbf{x}_{2}, \dots, \mathbf{x}_{T}$. The segmentation network is then trained using a  multi-loss function proposed in the next section. 

\subsection{Segmentation via Multi-Loss Formulation}

\par In histopathology, segmentation is majorly based on the structural aspects of the underlying images. Further, on many occasions, class imbalance is also inevitable because of patch-based analysis, which often introduces class dominance. Hence we propose a multi-loss function which is a combination of structural similarity (SS) \cite{b48} and focal loss (FL) \cite{b68}, which simultaneously caters to preserving structural importance and mitigating class imbalance.

\subsubsection{Structural Similarity Loss}

Structural similarity (SS) loss is designed to achieve a high positive linear correlation between the ground truth and the predicted segmentation masks.
Zhao \emph{et al.} \cite{b48} proposed it as a reweighted version of the cross-entropy (CE) loss. 
However, we modify it as the weighted absolute error between ground truth and predicted segmentation masks with the weights being the CE loss, as
\begin{align}\label{eq11}
    \mathcal{L}_{SS}(y_{nc}, \hat{y}_{nc}) &= \mathcal{L}_{CE} \cdot f_{nc} \cdot e_{nc}, \nonumber \\
    e_{nc} &= \Big|\frac{y_{nc}-\mu_{y_{nc}}+C_{1}}{\sigma_{y_{nc}}+C_{1}} - \frac{\hat{y}_{nc}-\mu_{\hat{y}_{nc}}+C_{1}}{\sigma_{\hat{y}_{nc}}+C_{1}}\Big|, \nonumber \\
    f_{nc} &= \mathbf{1}_{\{e_{nc} > \beta e_{\text{max}} \}}, \text{ and } \nonumber \\
    \mathcal{L}_{CE} &= -\frac{1}{{N}}\sum_{n=0}^{{N}-1}\sum_{c=0}^{{C}-1} y_{nc}\log \hat{y}_{nc},
\end{align}
where $\mu_{y_{nc}}$ and $\sigma_{y_{nc}}$ are the mean and standard deviation of the ground truth $y_{nc}$, respectively. $n$ and $c$ correspond to batch $N$ and channels $C$, respectively. $\hat{y}$ is the predicted segmentation mask and $C_1 = 0.01$ is an empirically set stability factor. The absolute error $e_{nc}$ measures the degree of linear correlation between two image patches. 
$e_{\text{max}}$ is the maximum value of $e_{nc}$, $\beta \in [0, 1)$ is a weight factor with $\beta = 0.1$ in practice, $\mathbf{1_{\{ . \}}}$ is the indicator function, and $\mathcal{L}_{CE}$ is the cross-entropy loss.
The structural similarity loss is expressed as
\begin{equation}\label{eq12}
    \mathcal{L}_{SS} = \frac{1}{M}\sum_{n=0}^{{N}-1}\sum_{c=0}^{{C}-1} \mathcal{L}_{SS} (y_{nc}, \hat{y}_{nc})
\end{equation}
where $M = \sum_{n=0}^{{N}-1}\sum_{c=0}^{{C}-1} f_{nc}$ is the number of hard examples. $f_{nc}$ is used to identify the pixels with significant absolute error between the predicted and ground truth segmentation masks for every class, and $\mathcal{L}_{CE}$ adds weighting to those pixels. 
Here, $\mathcal{L}_{CE}$ is the dynamic weighting factor varying over the iterations based on the prediction.
We empirically observe the effect of this modified structural similarity loss in boosting the segmentation performance in Section \ref{loss effects}. 

\subsubsection{Focal Loss}
Focal loss is shown to perform well in the presence of imbalanced datasets \cite{b49}, defined as:
\begin{equation}\label{eq13}
    \mathcal{L}_{FL} = -\frac{1}{{N}}\sum_{n=0}^{{N}-1}\sum_{c=0}^{{C}-1}(1-\hat{y}_{nc})^{\gamma}y_{nc}\log\hat{y}_{nc},
\end{equation}
where $N$ denotes the batch size, $C$ denotes the number of classes, and $y_{nc}$, $\hat{y}_{nc}$ are the ground truth and predicted values for any pixel corresponding to a class, respectively. $(1-\hat{y}_{nc})^{\gamma}$ acts as a weighting factor and takes care of the class imbalance. 
The value of $\gamma$ is set to $2.0$ empirically. The final loss function for supervised fine-tuning our method is a weighted combination of the structural similarity and the focal loss given by $\mathcal{L}_{SSFL} = \mathcal{L}_{SS} + \lambda \mathcal{L}_{FL}$, where $\lambda$ is a hyperparameter.

\section{Experiments and Results}\label{sec:experiments}

\subsection{Datasets}

For our experiments, we use three datasets, namely the Head and Neck Cancer, Gland segmentation in colon histology images (GlaS) \cite{b51}, and multi-organ nucleus segmentation (MoNuSeg) \cite{b50} datasets. The details of the datasets used are given in Table \ref{tab:datasets}. While the latter two are publicly available, the former is curated, annotated, and proposed by us for research community usage. The dataset will be shared for research purposes upon request. 

\begin{table}[h]
\caption{Details About the Proposed HN Cancer Dataset.}
\label{tab:hn_table}
\centering
\setlength{\tabcolsep}{3pt}
\begin{tabular}{c|c|c}
\hline
 & Attribute & Value \\ \hline
\multirow{5}{*}{\begin{tabular}[c]{@{}c@{}}Imaging\\ Details\end{tabular}} & Image Type & Whole tissue section \\
 & Optical Magnification & 10x \\
 & Imgaing System & Nikon Eclipse 80i microscope \\
 & Image Dimension & 1024 x 1280 \\
 & Cancer Size & 1 - 4 cm \\ \hline
\multirow{4}{*}{\begin{tabular}[c]{@{}c@{}}Dataset\\ Details\end{tabular}} & Cancer Classes & \begin{tabular}[c]{@{}c@{}}Malignant, non-malignant stroma \\ and non-malignant epithelium\end{tabular} \\ \cline{2-3} 
 & Malignant & SCC tumor cell islands \\ \cline{2-3} 
 & Non-malignant stroma & \begin{tabular}[c]{@{}c@{}}Fibroblasts, inflammatory cells, \\ endothelial cells, adipose tissue, \\ skeletal muscle\end{tabular} \\ \cline{2-3} 
 & Non-malignant epithelium & Benign squamous epithelium \\ \hline
\multirow{2}{*}{Statistics} & Annotated Images & 505 \\
 & Unannotated Images & 1057 \\ \hline
\end{tabular}
\end{table}

\subsubsection{Head and Neck Cancer Dataset} 
\label{subsec: HN cancer}
This dataset was collected with the approval of the Ethics Committee of the All India Institute of Medical Sciences (AIIMS), New Delhi, with the approval number IEC-58/04.01.2019.
A total of $163$ cases of head and neck squamous cell carcinoma (SCC) \cite{b65} were retrieved from the archives of the Department of Pathology, AIIMS. Tumor tissue had been fixed in 10\% neutral buffered formalin, routinely processed, and embedded into paraffin blocks. Representative H\&E stained sections lacking cutting artifacts were selected.
Images of whole tissue sections (and not scanned whole slide images) were captured at 100x magnification (10x objective lens, 10x eyepiece lens) using a digital camera attached to a microscope using a Nikon Eclipse 80i microscope with a camera and the \textit{NIS-Elements} image capture software.
Whole slide imaging (WSI) refers to the automated capture of high-resolution digitized images of histological slides by using a digital slide scanner. The WSI is generated by digital stitching together of individual images captured by camera on a robotic microscope. For computational pathology, such digitally scanned whole slide images are broken up into a number of smaller sub-images or patches, which then undergo further analysis \cite{b79, b80}. This essentially gives the same results as utilizing individually captured images. Most machine learning based classifiers are built on such sub-images extracted from WSI rather than the entire WSI.

The chief advantage of using WSI is that a skilled pathologist is not required to capture images of histological slides, which is a time and labor-intensive task. With WSI, a large number of sub-images can be generated rapidly, resulting in time saving and complete coverage of the entire area in a tissue section. However, use of WSIs in computational pathology has its own challenges. The heavy file size and weight of WSI impede data storage and transfer, and also affect computation time. Annotation of WSI often involves identifications of regions of interest (ROI) with cancer associated patterns, rather than direct annotation of cancer and non-cancer cells, as reported by Halicek et al. \cite{b81} and Bulten et al. \cite{b82}.

The methodology described here for capturing of images is the standard technique for capturing histopathological images for any image analysis application. Rahman et al.\cite{b83} similarly captured images using a Leica microscope with a camera using 10x objective lens and 10x eyepiece lens for one set of images, as we have done, and using 40x objective lens and 10x eyepiece lens for a second set of images. They proposed that the first set of images may be used for feature extraction, segmentation and classification purposes, and for establishing automated decision support systems. Krishnan et al. \cite{b84} similarly captured images of histological sections using a Zeiss Microscope under 10x objective in their analysis of oral submucous fibrosis biopsies \cite{b80}. The Breakhis Breast Cancer Histopathological Database is one of the largest breast cancer datasets \cite{b85,b86}. It utilizes microscopic images captured in a similar manner at various magnifications including 10x, 100x, 200x, and 400x \cite{b81,b82}.

Each captured image had dimensions of 1024 x 1280 pixels.
The size of the majority of head and neck cancers ranges from 1 to 4 cm. Each tumor was sampled such that one section was taken per cm of a tumor, with each section having an area of approximately 1.5 x 1.5 cm (a 10x field has a diameter of 3mm). Thus, for adequate representation of each tumor and intervening non-tumor areas, a minimum of 4 images were captured, after avoiding areas of necrosis. 
A team of trained pathologists performed manual annotation of the captured images. Images were annotated using the online image annotation tool called \textit{labelbox} \footnote{\url{https://labelbox.com/product/annotate/image/}} into three classes: malignant, non-malignant stroma, and non-malignant epithelium. 
All SCC tumor cell islands were marked as malignant. 
Non-malignant stroma included fibroblasts, inflammatory cells, endothelial cells, adipose tissue, and skeletal muscle.
Immune cells were included in the non-malignant stroma class. 
Care was taken not to include necrotic foci in the captured images. 
Non-malignant epithelium included all benign squamous epithelium adjacent to the tumor or from resection margins. 
All the tissue present in an image was annotated. 
Diagnosis of cancer requires distinction of cancer areas from the non-malignant epithelium and demonstration of invasion by cancer into the non-malignant stroma. 
Thus, delineating the three classes would aid a pathologist in identifying invasion by cancer cells. 
We use images from confirmed cases of cancers that have been surgically removed completely, and there is no ambiguity in the diagnosis, as we have ample tissue to study under the microscope. 
There are $1562$ images in the collected dataset, of which $505$ are annotated. 
The dataset details have been consolidated into Table \ref{tab:hn_table}.

\subsubsection{Public Datasets}
We analyze two publicly available datasets: GlaS \cite{b51} and MoNuSeg \cite{b50}. GlaS contains $165$ images from $16$ H\&E stained sections of stage T3/T4 colorectal adenocarcinoma, showing notable inter-subject variability in stain distribution and tissue architecture. Images are mostly $775 \times 522$ in resolution. $85$ images are used for training and $80$ for testing. We use the training set for self-supervised pre-training and split the $80$ test images into train-test sets for segmentation. $52$ visual fields from malignant and benign regions were selected for diverse tissue architectures. Pathologists annotated glandular boundaries, categorizing regions into malignant and benign classes.

\begin{figure}[!t]
\centerline{\includegraphics[width=\columnwidth,trim=1.5cm 0.5cm 1.5cm 0.5cm,clip]{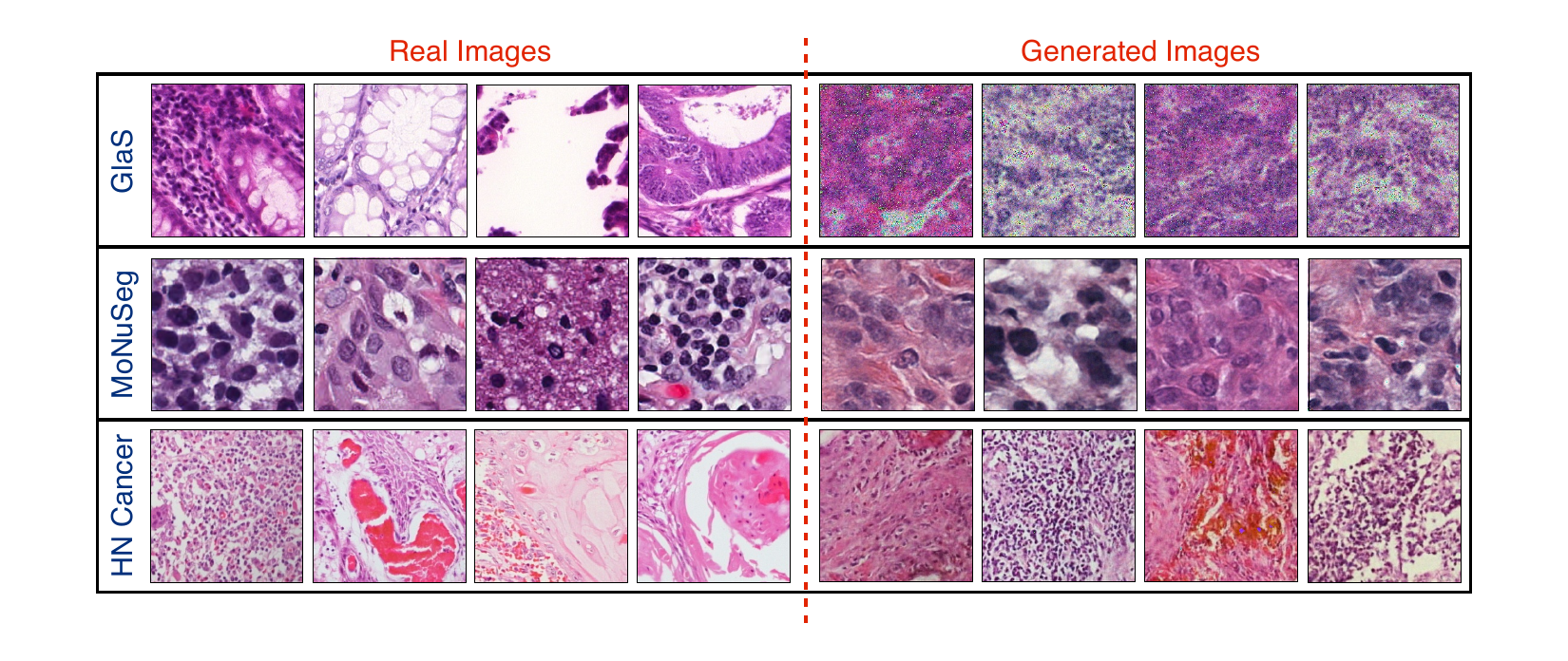}}
\caption{Sample real and generated patches using diffusion on three datasets: GlaS \cite{b50}, MoNuSeg \cite{b51}, and HN cancer (ours). The first four images in each row represent real image patches, and the last four images represent the generated image patches.}
\label{fig:samples_datasets}
\end{figure}

The MoNuSeg dataset consists of H\&E stained images at 40x magnification. For nuclear appearance diversity, one image per patient was chosen and cropped into $1000 \times 1000$ sub-images dense in nuclei. Annotated classes cover epithelial and stromal nuclei, resolving overlaps in classes by assigning pixels to the largest nucleus. This dataset contains $37$ training and $14$ testing images. Initially, three classes are annotated: nucleus boundary, within-nucleus (foreground), and outside-nuclei (background). However, our study combines the nucleus boundary and nucleus into one class. Our pre-training stage involves all $37$ training images, while the remaining 14 are used for segmentation. 
Sample images from all the evaluation datasets are shown in Fig. \ref{fig:samples_datasets}.

The GlaS and MoNuSeg datasets use fixed training and testing sets. The training set images serve as unlabeled data for self-supervised pre-training, while the testing set images provide labeled data for segmentation. 
This guarantees that pre-training and fine-tuning stages use mutually exclusive image sets. 
Labeled images are further divided into train and test subsets, with the latter used only for performance assessment.
It was ensured that there is patient-level separation between the training and testing sets. 
The HN cancer dataset also comprises separate labeled and unlabeled images. 
Patches of 256 × 256 (stride of 64 for GlaS and MoNuSeg, 256 for HN cancer) are extracted for training and evaluation. Dataset details are in Table \ref{tab:datasets}. 
In addition, we also conduct experiments on MoNuSeg and GlaS where the pre-training and downstream training data are the same. 
In this case, the pre-training part uses the unlabelled images from the official training set and the downstream stage uses the same images but with annotations.

\begin{table}
\caption{Number of Images and Patches in Each Dataset. The Numbers in Brackets (in red) Indicate the Patches.}
\label{table1}
\centering
\setlength{\tabcolsep}{3pt}
\begin{tabular}{p{35pt}p{60pt}p{40pt}p{40pt}}
\hline
Dataset& 
Unlabeled Images& 
\multicolumn{2}{p{80pt}}{Labeled Images} \\
\cmidrule(lr){3-4}
 & & Train & Test\\
\cmidrule(lr){1-1}
\cmidrule(lr){2-2}
\cmidrule(lr){3-3}
\cmidrule(lr){4-4}
GlaS& 
85 (\textcolor{red}{3681})& 
64 (\textcolor{red}{2685}) &  16 (\textcolor{red}{720})\\
MoNuSeg& 
37 (\textcolor{red}{5328})& 
11 (\textcolor{red}{1584}) &  3 (\textcolor{red}{432})\\
HN cancer& 
1057 (\textcolor{red}{21140})& 
404 (\textcolor{red}{8080}) &  101 (\textcolor{red}{2020})\\
\hline
\end{tabular}
\label{tab:datasets}
\end{table}

\begin{table}[t]
\caption{Generation Performance (FID Scores) of Self-Supervised Pre-Training on Three Datasets: GlaS, MoNuSeg, and HN Cancer and the Combination of all Datasets.}
\begin{center}
\setlength{\tabcolsep}{3pt}
\begin{tabular}{p{45pt}p{30pt}p{40pt}p{40pt}p{40pt}}
\hline
Dataset & {\centering GlaS} & {\centering MoNuSeg} & {\centering HN cancer} & Combined\\
\hline
FID Score & {\centering 82.44} & {\centering 15.32} & {\centering 6.80} & {\centering 7.63}\\
\hline
\end{tabular}
\label{tab:generation_perf}
\end{center}
\end{table}

\begin{table*}[]
\caption{Performance Evaluation of the Proposed Method and Baseline Methods on GlaS, MoNuSeg, and HN Cancer Datasets. AJI, IoU, HD, and F1 indicate Aggregated Jaccard Index, Intersection over Union, Hausdorff Distance, and F1-score respectively. AJI and IoU are the same for GlaS and MoNuSeg since there is only one nucleus class.}
\setlength{\tabcolsep}{6pt}
\begin{center}
\begin{tabular}{lcccccccccccc}
\hline
\multicolumn{1}{c|}{\multirow{2}{*}{Method}} & \multicolumn{4}{c|}{GlaS} & \multicolumn{4}{c|}{MoNuSeg} & \multicolumn{4}{c}{HN Cancer} \\ 
\multicolumn{1}{c|}{} & AJI & IoU & HD & \multicolumn{1}{c|}{F1} & AJI & IoU & HD & \multicolumn{1}{c|}{F1} & AJI & IoU & HD & F1 \\ \hline
\multicolumn{13}{c}{Supervised Methods} \\ \hline
\multicolumn{1}{l|}{UNet \cite{b2}} & 0.7565 & 0.7565 & 7.6214 & \multicolumn{1}{c|}{0.8314} & 0.6042 & 0.6042 & 7.5719 & \multicolumn{1}{c|}{0.7515} & 0.6912 & 0.6067 & 6.3234 & 0.7280 \\
\multicolumn{1}{l|}{Attention UNet \cite{b66}} & 0.7924 & 0.7924 & 7.0954 & \multicolumn{1}{c|}{0.8614} & 0.6221 & 0.6221 & 7.5729 & \multicolumn{1}{c|}{0.7657} & 0.7648 & 0.7772 & 5.0056 & 0.8091 \\ 
\multicolumn{1}{l|}{FCT \cite{b78}} & 0.7991 & 0.7991 & 6.9557 & \multicolumn{1}{c|}{0.8620} & 0.5880 & 0.5880 & 7.6901 & \multicolumn{1}{c|}{0.7356} & 0.7395 & 0.7362 & 5.4375 & 0.7740 \\ \hline
\multicolumn{13}{c}{DDPM-Based Supervised Methods} \\ \hline
\multicolumn{1}{l|}{Baranchuk et. al \cite{b74}} & 0.6544 & 0.6544 & 8.9907 & \multicolumn{1}{c|}{0.7593} & 0.5982 & 0.5982 & 7.5714 & \multicolumn{1}{c|}{0.7456} & 0.6318 & 0.5746 & 7.4952 & 0.6275 \\
\multicolumn{1}{l|}{Wolleb et. al \cite{b73}} & 0.7837 & 0.7837 & 7.1489 & \multicolumn{1}{c|}{0.8626} & 0.6055 & 0.6055 & 7.6902 & \multicolumn{1}{c|}{0.7521} & 0.6064 & 0.3802 & 9.5552 & 0.4017 \\
\multicolumn{1}{l|}{CIMD \cite{b72}} & 0.5694 & 0.5694 & 8.7733 & \multicolumn{1}{c|}{0.6928} & 0.5938 & 0.5938 & 7.8550 & \multicolumn{1}{c|}{0.7433} & 0.5915 & 0.4086 & 9.2612 & 0.4632 \\
\multicolumn{1}{l|}{MedSegDiff \cite{b21}} & 0.7788 & 0.7788 & 7.5456 & \multicolumn{1}{c|}{0.8521} & 0.5734 & 0.5734 & 7.8535 & \multicolumn{1}{c|}{0.7270} & 0.6272 & 0.3904 & 9.4729 & 0.4115 \\ \hline
\multicolumn{13}{c}{Self-Supervised Pretext Tasks} \\ \hline
\multicolumn{1}{l|}{VAE \cite{b18}} & 0.7372 & 0.7372 & 7.4800 & \multicolumn{1}{c|}{0.8102} & 0.5978 & 0.5978 & 7.7309 & \multicolumn{1}{c|}{0.7459} & 0.7481 & 0.7572 & 5.2894 & 0.7886 \\
\multicolumn{1}{l|}{Context Rest. \cite{b4}} & 0.8154 & 0.8154 & 6.8289 & \multicolumn{1}{c|}{0.8787} & 0.6284 & 0.6284 & 7.4162 & \multicolumn{1}{c|}{0.7697} & 0.7465 & 0.7853 & 4.8376 & 0.8148 \\
\multicolumn{1}{l|}{Contrastive \cite{b40}} & 0.7614 & 0.7614 & 7.5377 & \multicolumn{1}{c|}{0.8329} & 0.5979 & 0.5979 & 7.7223 & \multicolumn{1}{c|}{0.7461} & 0.7343 & 0.7564 & 5.2289 & 0.7851 \\
\multicolumn{1}{l|}{CS-CO \cite{b41}} & 0.8025 & 0.8025 & 7.0016 & \multicolumn{1}{c|}{0.8699} & 0.5342 & 0.5342 & 8.0614 & \multicolumn{1}{c|}{0.6950} & 0.7479 & 0.7477 & 5.1751 & 0.7813 \\
\multicolumn{1}{l|}{DIM \cite{b52}} & 0.8083 & 0.8083 & 6.8582 & \multicolumn{1}{c|}{0.8700} & 0.6243 & 0.6243 & 7.6320 & \multicolumn{1}{c|}{0.7671} & 0.7572 & 0.7779 & 5.0219 & 0.8084 \\
\multicolumn{1}{l|}{Inpainting \cite{b8}} & 0.7787 & 0.7787 & 7.1050 & \multicolumn{1}{c|}{0.8522} & 0.6011 & 0.6011 & 7.7126 & \multicolumn{1}{c|}{0.7471} & 0.7419 & 0.7551 & 5.2598 & 0.7835 \\
\multicolumn{1}{l|}{CycleGAN \cite{b69}} & 0.8053 & 0.8053 & 6.8834 & \multicolumn{1}{c|}{0.8694} & 0.6182 & 0.6182 & 7.4654 & \multicolumn{1}{c|}{0.7619} & 0.7690 & 0.7240 & 5.4719 & 0.7587 \\ \hline
\multicolumn{1}{l|}{Ours} & \textbf{0.8470} & \textbf{0.8470} & \textbf{6.3517} & \multicolumn{1}{c|}{\textbf{0.9026}} & \textbf{0.6545} & \textbf{0.6545} & \textbf{7.3261} & \multicolumn{1}{c|}{\textbf{0.7895}} & \textbf{0.7913} & \textbf{0.8105} & \textbf{4.6855} & \textbf{0.8413} \\ \hline
\end{tabular}
\label{tab:main_table}
\end{center}
\end{table*}

\begin{table}[]
\caption{Performance Evaluation on GlaS and MoNuSeg Datasets Using the Same Data for Pre-Training and Downstream Training (Following Official Train-Test Splits).}
\setlength{\tabcolsep}{3pt}
\begin{center}
\begin{tabular}{lcccccc}
\hline
\multicolumn{1}{c|}{\multirow{2}{*}{Method}} & \multicolumn{3}{c|}{GlaS} & \multicolumn{3}{c}{MoNuSeg}\\ 
\multicolumn{1}{c|}{} & AJI & HD & \multicolumn{1}{c|}{F1} & AJI & HD & F1 \\ \hline
\multicolumn{1}{l|}{UNet \cite{b2}} & 0.7397 & 8.0446 & \multicolumn{1}{c|}{0.8215} & 0.6403 & 7.6432 & 0.7786 \\
\multicolumn{1}{l|}{Attention UNet \cite{b66}} & 0.7607 & 7.3252 & \multicolumn{1}{c|}{0.8276} & 0.6717 & 7.4051 & 0.7997 \\ 
\multicolumn{1}{l|}{FCT \cite{b78}} & 0.7525 & 7.4942 & \multicolumn{1}{c|}{0.8252} & 0.6502 & 7.3963 & 0.7848 \\ \hline
\multicolumn{1}{l|}{Baranchuk et. al \cite{b74}} & 0.6251 & 9.4429 & \multicolumn{1}{c|}{0.7375} & 0.6547 & 7.4312 & 0.7894 \\
\multicolumn{1}{l|}{Wolleb et. al \cite{b73}} & 0.6670 & 8.1465 & \multicolumn{1}{c|}{0.7376} & 0.6732 & 7.2712 & 0.8030 \\
\multicolumn{1}{l|}{CIMD \cite{b72}} & 0.5059 & 10.8664 & \multicolumn{1}{c|}{0.6250} & 0.5371 & 8.1764 & 0.6925 \\
\multicolumn{1}{l|}{MedSegDiff \cite{b21}} & 0.5637 & 10.0343 & \multicolumn{1}{c|}{0.6459} & 0.4663 & 9.0639 & 0.6685 \\ \hline
\multicolumn{1}{l|}{VAE \cite{b18}} & 0.7583 & 7.6375 & \multicolumn{1}{c|}{0.8296} & 0.6733 & 7.3184 & 0.8024 \\
\multicolumn{1}{l|}{Context Rest. \cite{b4}} & 0.7661 & 7.3518 & \multicolumn{1}{c|}{0.8336} & 0.6661 & 7.4881 & 0.7924 \\
\multicolumn{1}{l|}{Contrastive \cite{b40}} & 0.7736 & 7.3983 & \multicolumn{1}{c|}{0.8416} & 0.6460 & 7.5402 & 0.7802 \\
\multicolumn{1}{l|}{CS-CO \cite{b41}} & 0.7664 & 7.4101 & \multicolumn{1}{c|}{0.8387} & 0.6051 & 7.5442 & 0.7525 \\
\multicolumn{1}{l|}{DIM \cite{b52}} & 0.7755 & 7.3522 & \multicolumn{1}{c|}{0.8424} & 0.6736 & 7.3109 & 0.8032 \\
\multicolumn{1}{l|}{Inpainting \cite{b8}} & 0.7687 & 7.4038 & \multicolumn{1}{c|}{0.8377} & 0.6587 & 7.3561 & 0.7916 \\
\multicolumn{1}{l|}{CycleGAN \cite{b69}} & 0.7766 & 7.3944 & \multicolumn{1}{c|}{0.8478} & 0.6650 & 7.3699 & 0.7970 \\ \hline
\multicolumn{1}{l|}{Ours} & \textbf{0.8034} & \textbf{6.9270} & \multicolumn{1}{c|}{\textbf{0.8652}} & \textbf{0.6901} & \textbf{7.1157} & \textbf{0.8147} \\ \hline
\end{tabular}
\label{tab:official_same}
\end{center}
\end{table}

\begin{table}[t]
\caption{Performance Evaluation on MoNuSeg with Separated Nucleus and Boundary Classes Using the Standard (Official) Train-Test Split.}
\setlength{\tabcolsep}{6pt}
\begin{center}
\begin{tabular}{lcccccccccccc}
\hline
\multicolumn{1}{c|}{Method} & AJI & IoU & HD & F1 \\ \hline
\multicolumn{1}{l|}{UNet \cite{b2}} & 0.5152 & 0.3647 & 6.2504 & 0.4766 \\
\multicolumn{1}{l|}{Attention UNet \cite{b66}} & 0.5244 & 0.3775 & 6.2218 & 0.4905 \\ 
\multicolumn{1}{l|}{FCT \cite{b78}} & 0.5296 & 0.3824 & 6.2117 & 0.4946 \\ \hline
\multicolumn{1}{l|}{Baranchuk et. al \cite{b74}} & 0.3583 & 0.3207 & 6.2961 & 0.4033 \\
\multicolumn{1}{l|}{Wolleb et. al \cite{b73}} & 0.4973 & 0.3569 & 6.2966 & 0.4518 \\
\multicolumn{1}{l|}{CIMD \cite{b72}} & 0.3202 & 0.2361 & 8.5954 & 0.3102 \\
\multicolumn{1}{l|}{MedSegDiff \cite{b21}} & 0.2927 & 0.2336 & 7.3343 & 0.2850 \\ \hline
\multicolumn{1}{l|}{VAE \cite{b18}} & 0.5310 & 0.3833 & 6.2088 & 0.4971 \\
\multicolumn{1}{l|}{Context Rest. \cite{b4}} & 0.5332 & 0.3862 & 6.1957 & 0.4996 \\
\multicolumn{1}{l|}{Contrastive \cite{b40}} & 0.5015 & 0.3610 & 6.4057 & 0.4738 \\
\multicolumn{1}{l|}{CS-CO \cite{b41}} & 0.4555 & 0.3174 & 6.2210 & 0.4251 \\
\multicolumn{1}{l|}{DIM \cite{b52}} & 0.5290 & 0.3808 & 6.2530 & 0.4931 \\
\multicolumn{1}{l|}{Inpainting \cite{b8}} & 0.5141 & 0.3688 & 6.2577 & 0.4809 \\
\multicolumn{1}{l|}{CycleGAN \cite{b69}} & 0.5236 & 0.3764 & 6.2637 & 0.4905 \\ \hline
\multicolumn{1}{l|}{Ours} & \textbf{0.5583} & \textbf{0.4039} & \textbf{6.0760} & \textbf{0.5146} \\ \hline
\end{tabular}
\label{tab:monuseg_official}
\end{center}
\end{table}

\begin{table}[t]
\caption{Performance Evaluation on MoNuSeg with Separated Nucleus and Boundary Classes Using Mutually exclusive Pre-Training and Downstream Train Sets.}
\setlength{\tabcolsep}{6pt}
\begin{center}
\begin{tabular}{lcccccccccccc}
\hline
\multicolumn{1}{c|}{Method} & AJI & IoU & HD & F1 \\ \hline
\multicolumn{1}{l|}{UNet \cite{b2}} & 0.5324 & 0.3763 & 6.0837 & 0.4833 \\
\multicolumn{1}{l|}{Attention UNet \cite{b66}} & 0.5357 & 0.3809 & 6.0412 & 0.4857 \\ 
\multicolumn{1}{l|}{FCT \cite{b78}} & 0.5521 & 0.3912 & 6.0097 & 0.4955 \\ \hline
\multicolumn{1}{l|}{Baranchuk et. al \cite{b74}} & 0.3702 & 0.3303 & 6.1667 & 0.3982 \\
\multicolumn{1}{l|}{Wolleb et. al \cite{b73}} & 0.4530 & 0.3512 & 6.2516 & 0.4574 \\
\multicolumn{1}{l|}{CIMD \cite{b72}} & 0.3588 & 0.3125 & 6.6065 & 0.3843 \\
\multicolumn{1}{l|}{MedSegDiff \cite{b21}} & 0.3384 & 0.2898 & 7.1564 & 0.3417 \\ \hline
\multicolumn{1}{l|}{VAE \cite{b18}} & 0.5429 & 0.3838 & 6.0715 & 0.4912 \\
\multicolumn{1}{l|}{Context Rest. \cite{b4}} & 0.5518 & 0.3961 & 6.0147 & 0.5038 \\
\multicolumn{1}{l|}{Contrastive \cite{b40}} & 0.4601 & 0.3702 & 6.0299 & 0.4691 \\
\multicolumn{1}{l|}{CS-CO \cite{b41}} & 0.3363 & 0.2880 & 6.4868 & 0.3653 \\
\multicolumn{1}{l|}{DIM \cite{b52}} & 0.5535 & 0.3976 & 6.0152 & 0.5024 \\
\multicolumn{1}{l|}{Inpainting \cite{b8}} & 0.5419 & 0.3805 & 6.0321 & 0.4864 \\
\multicolumn{1}{l|}{CycleGAN \cite{b69}} & 0.5492 & 0.3939 & 6.0652 & 0.5037 \\ \hline
\multicolumn{1}{l|}{Ours} & \textbf{0.5764} & \textbf{0.4067} & \textbf{5.9072} & \textbf{0.5135} \\ \hline
\end{tabular}
\label{tab:monuseg_split}
\end{center}
\end{table}

\begin{table*}[htbp]
\caption{Mean and Standard Deviation on GlaS, MoNuSeg, and HN Cancer for all the Metrics for our Method.}
\begin{center}
\setlength{\tabcolsep}{3pt}
\begin{tabular}{p{50pt}p{85pt}p{65pt}p{65pt}p{65pt}}
\hline
Dataset & AJI & IoU & HD & F1\\
\hline
GlaS & 0.8455 $\pm$ 0.0013 & 0.8455 $\pm$ 0.0013 & 6.3112 $\pm$ 0.0444 & 0.9014 $\pm$ 0.0009\\
MoNuSeg & 0.6463 $\pm$ 0.0062 & 0.6463 $\pm$ 0.0062 & 7.3511 $\pm$ 0.0200 & 0.7829 $\pm$ 0.0050\\
HN cancer & 0.7851 $\pm$ 0.0055 & 0.8050 $\pm$ 0.0039 & 4.7113 $\pm$ 0.0199 & 0.8353 $\pm$ 0.0044\\
\hline
\end{tabular}
\label{tab:mean_std}
\end{center}
\end{table*}

\subsection{Baselines for Comparison }\label{sec:comparisons}

\begin{enumerate}[wide, labelindent=0pt]
\item \textit{Supervised Methods:} We compare our framework with three fully supervised benchmarks: (a) U-Net \cite{b2} - a standard model for medical image segmentation, and (b) Attention U-Net \cite{b66} with random initialization: a U-Net incorporating an attention mechanism designed for CT abdominal image segmentation. (3) Fully convolutional transformer \cite{b78}: a transformer to segment medical images with multiple modalities.
\item \textit{DDPM-Based Methods:} We compare with DDPM-based supervised methods: (1) Baranchuk et al. \cite{b74}: A label efficient training strategy using DDPM. (2) Wolleb et al. \cite{b73} introduced a segmentation method using an ensemble of DDPM mask predictions. (3) CIMD \cite{b72}: a DDPM method that learns a distribution over segmentation masks with minimal training overhead. (4) MedSegDiff \cite{b21}: a conditional encoding scheme to perform segmentation using a DDPM.   
\item \textit{Pretext Tasks:} The remaining baselines adopt diverse pretext tasks for self-supervision: (1) VAE \cite{b18}: A U-Net-based variational autoencoder pre-trained and downstreamed for segmentation. (2) Context restoration \cite{b4}: Self-supervised learning through context restoration, targeted at medical image analysis. (3) Contrastive learning \cite{b40}: Leveraging self-supervised contrastive learning for acquiring image representations. (4) CS-CO \cite{b41}: A histopathological image-specific self-supervised method that integrates predictive and contrastive learning techniques with novel image augmentations. (5) Deep InfoMax (DIM) \cite{b52}: Unsupervised representation learning by maximizing mutual information between input and output. (6) Inpainting \cite{b8}: U-Net model trained using image inpainting as a pretext task. (7) CycleGAN \cite{b69}: An image-to-image GAN trained with cycle consistency.
\end{enumerate} 
All methods are trained until loss convergence, followed by fine-tuning for segmentation. 
For the supervised methods, we do not make use of the unlabeled images since there is no pretraining involved. To ensure fairness in comparisons, all the methods use the same UNet backbone.

\subsection{Implementation Details}
We use the attention-based U-Net as the encoder-decoder network for pre-training and fine-tuning. 
The time stamps are drawn randomly based on a uniform distribution between $0$ and $T$ and then input to the corresponding embedding layer for the network during pre-training (refer Fig \ref{fig:framework}). 
We train the network for $100$ epochs with a learning rate of $0.0001$ using Adam optimizer and a batch size of $8$. 
For the downstream segmentation, we initialize the U-Net, except for the last few layers, with the pre-trained weights of the pretext task and then fine-tune the entire network end-to-end with the Adam optimizer for $150$ epochs. 
The batch size is $8$, and the learning rate is $0.0001$. 
The multi-loss function's regularization scaling parameter $\lambda$ is set to $1.0$. 
We use random horizontal and vertical flips, color jittering, and Gaussian blur as augmentations. 
All the comparisons are with U-Net-based networks except for CS-CO, a ResNet-based network. Optimal hyperparameters are chosen based on the validation performance (validation set is chosen as 10\% of the training set) across all the methods. 
We use the Aggregated Jaccard Index (AJI), Intersection over Union (IoU), Hausdorff Distance (HD), and F1-score as the evaluation metrics. 

\begin{figure*}[!ht]
\centering
\includegraphics[width=0.8\textwidth]{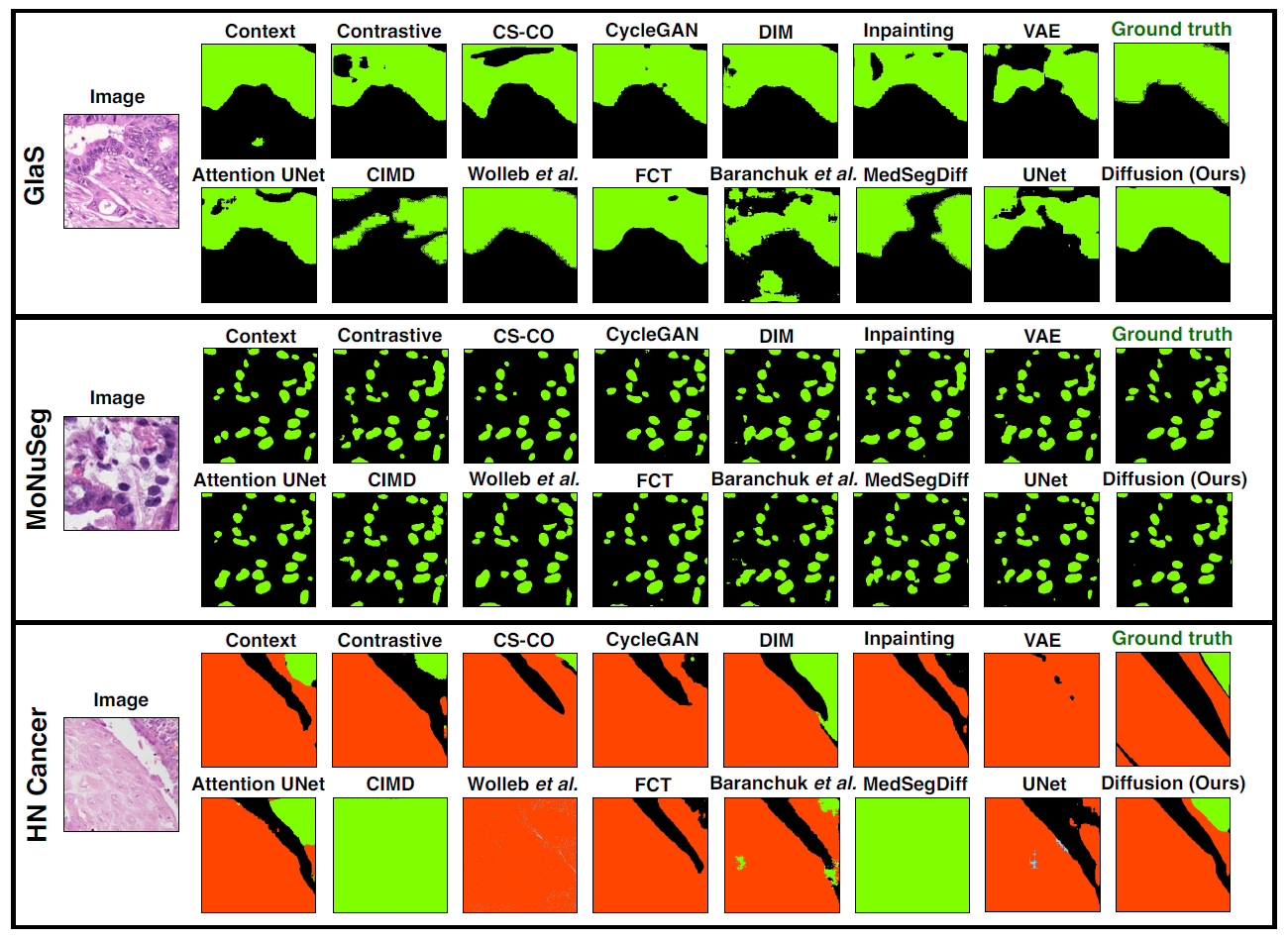}
\caption{Qualitative Results of the proposed method (diffusion) along with other pretext tasks (top row): Context Restoration \cite{b4}, Contrastive learning \cite{b40}, CS-CO \cite{b41}, CycleGAN \cite{b69}, DIM \cite{b52}, Inpainting \cite{b8} and VAE \cite{b18}. The bottom row contains fully supervised and DDPM-based methods: Attention UNet \cite{b66}, CIMD \cite{b72}, Wolleb et al. \cite{b73}, FCT \cite{b78}, Baranchuk et al. \cite{b74}, MedSegDiff \cite{b21}, UNet \cite{b2} and Diffusion (Ours).}
\label{fig:comparison_methods}
\end{figure*}

\subsection{Quantitative and Qualitative Results}
We demonstrate the effect of self-supervision tasks by transferring the learned representations to the histopathological image segmentation task.
Here, we compare our proposed diffusion-based self-supervision against other existing self-supervision tasks like context restoration \cite{b4}, contrastive learning \cite{b40}, and CS-CO \cite{b41}. 
The self-supervised approaches \cite{b40}, \cite{b41} are pathology specific, whereas \cite{b4} is pathology agnostic but still related to medical image analysis.
We also compare our method with supervised methods like attention UNet \cite{b66} and FCT \cite{b78}, and DDPM-based methods \cite{b74,b73,b72,b21}.
Moreover, we compare our approach with various methods as described in Section \ref{sec:comparisons}. 
Our method shares certain similarities with \cite{b44, b46, b47}, but these methods are mainly focused on generating high-quality images and learning good representations but are not tuned specifically for segmentation. 
\cite{b44} and \cite{b46} deal with DDPMs trained on natural images, while \cite{b47} focuses on synthesis of histopathology images.

Fig. \ref{fig:samples_datasets} shows examples of generated images from learning the self-supervised pretext task using diffusion. 
The generation performance of the diffusion model is captured using a popular metric, Frechet Inception Distance (FID) \cite{b67}, which is measured between images from a dataset and a set of generated images.
We use $1000$ real and generated images from each dataset to compute FID scores, which are shown in Table \ref{tab:generation_perf}.
One can observe that the FID scores are lower when the datasets contain more samples, indicating that the generation performance increases with the number of training patches. 
However, while combining all datasets, the FID score is observed to be slightly higher than that of HN cancer. This is because the distribution to be learned is more complex when domains are combined.
This can also be qualitatively understood in Fig. \ref{fig:samples_datasets}, where the generated images of GlaS, containing lower unlabeled images (see Table \ref{table1}), are noisy.
Moreover, we note that good generation performance also translates to good segmentation performance on all datasets. 
For the pre-training stage, we use unannotated images to train the DDPM using samples from $\mathcal{S}_{\text{pr}}$. 
We conduct two sets of experiments (with different training data settings) as follows:

\textit{Case I ($\mathcal{S}_{\text{pr}} \cap \mathcal{S}_{\text{tr}} = \phi$)}: 
Here, the pre-training and downstream training data are mutually exclusive. For downstream segmentation, the pre-trained UNet is further trained using annotated labels from $\mathcal{S}_{\text{tr}}$ (in this case $\mathcal{S}_{\text{tr}}$ is sampled from part of the official test sets). Table \ref{tab:main_table} shows the evaluation performance of all methods on all three segmentation datasets. 
We observe an improvement in AJI of at least $3.0 - 3.5$\% for GlaS, $2.5 - 3.0$\% for MoNuSeg, and $2.0 - 2.5$\% for HN cancer datasets.  
Moreover, we include performances of our approach on all three datasets over multiple runs in Table \ref{tab:mean_std} to demonstrate the stability in training.
Finally, Fig. \ref{fig:comparison_methods} demonstrates the superior qualitative performance of the proposed method over other self-supervised methods. 
To ensure fair comparisons, the best performances across multiple weight initializations are reported for all methods. 

\textit{Case II ($\mathcal{S}_{\text{pr}} = \mathcal{S}_{\text{tr}}$)}: In this case, the pre-training and downstream training use the same data (the corresponding ground truth labels are used only in the downstream training stage).
In the case of MoNuSeg and GlaS, training and testing data follow the official splits, and the corresponding results are presented in Table \ref{tab:official_same}.
We observe that the performance in the case of MoNuSeg is higher while using the official train-test split, since the amount of training patches used for downstream training is significantly higher.
However, in the case of GlaS, there is a slight drop observed in performance for all the methods due to a significant increase in the number of images for inference (in this case, the inference is done on the official test sets).

In addition, we evaluate the performance of all methods on MoNuSeg where the nucleus and its boundary are separated into two classes.
We observe from Tables \ref{tab:monuseg_official} and \ref{tab:monuseg_split} that our approach outperforms all other methods in both data split settings, validating its efficacy in segmentation.
Our approach achieves superior segmentation performance over other self-supervised methods on all the metrics for all three datasets under both training scenarios. 
Moreover, we observe that our method can outperform even DDPM-based models, since the existing methods require an ensemble of segmentation masks to be predicted, often inducing variance and randomness in the predictions.

\begin{figure}[t]
    \centering
    \begin{tabular}{cc}
        \includegraphics[width=0.48\columnwidth]{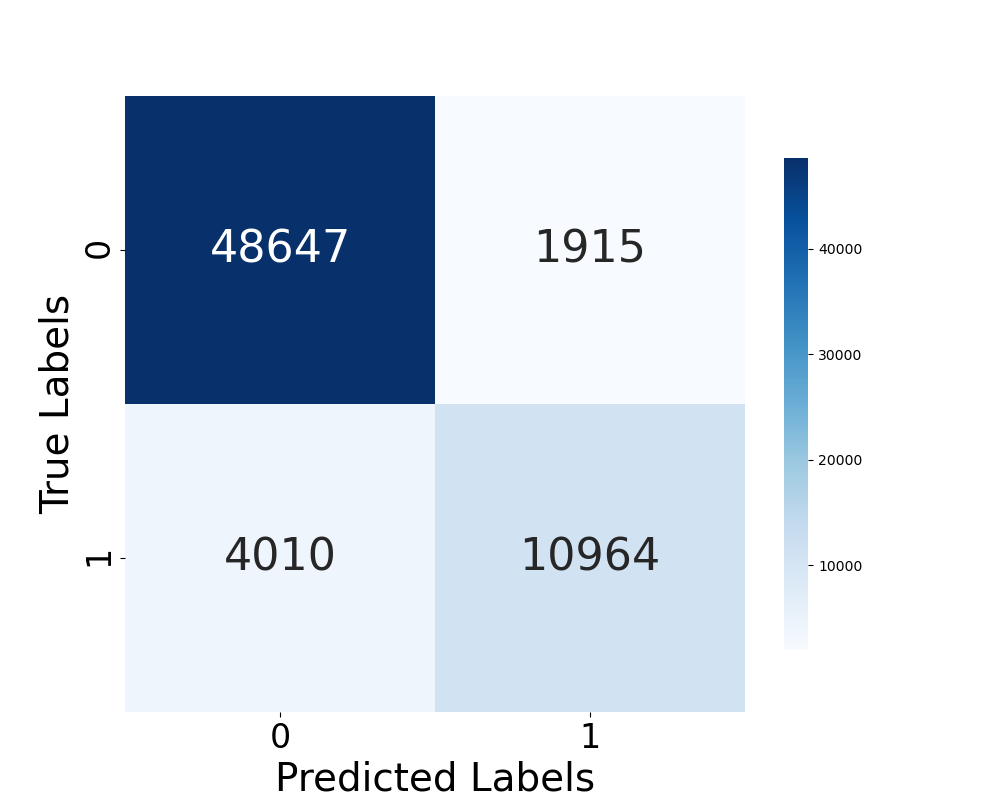} &
        \includegraphics[width=0.48\columnwidth]{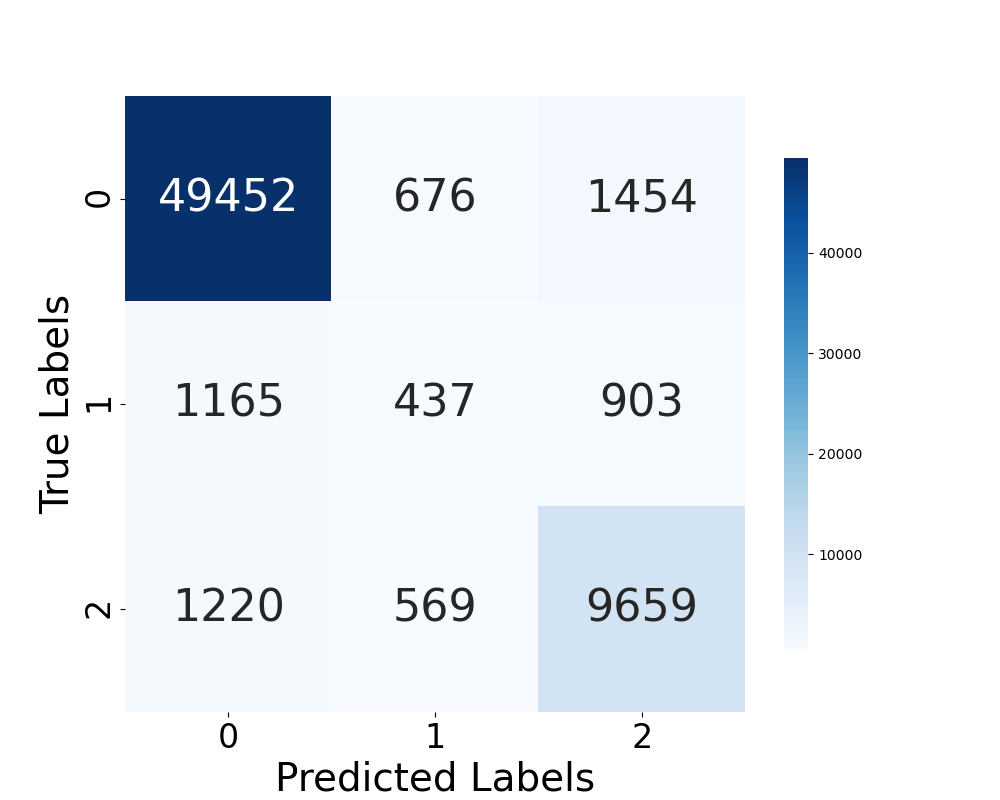} \\
        (a) & (b) \\
    \end{tabular}
    \caption{Confusion matrices for MoNuSeg (a) with the boundary and nucleus merged (with class 1 being nucleus), and (b) with boundary separated (here, class 0, 1 and 2 correspond to background, boundary and nucleus respectively).}
    \label{fig:confusion_matrices}
\end{figure}

We also pre-train a diffusion model using a combination of unannotated images from all three datasets and then learn separate segmentation models for each dataset. 
From Table \ref{tab:combined_datasets}, we notice the segmentation performance on all three datasets is comparable to the case where both stages of our framework are trained on a single dataset.
In the case of HN cancer, we observe a slight improvement in one of the metrics (AJI) since the combined training data has about $70\%$ training samples coming from the HN cancer dataset.
In the case of GlaS, there is a slight improvement in performance in all metrics.
However, in the case of MoNuSeg, there is a slight improvement in only HD.
We also observe that in the case of MoNuSeg, while going from two to three classes, there is a slight drop in performance. When dealing with two classes, the nucleus boundary can either be included in the foreground or background class depending on the cells being analyzed. Upon observing the images in the dataset, we notice that epithelial and stromal cells are present. Since the H\&E staining characteristics of nuclear boundary for both these cell types differs, the boundary would not be uniformly segmented into either the foreground or background classes. In the three-class case, this causes some of the boundary regions to either be classified as foreground or background, leading to more misclassifications (this can be observed from the confusion matrices shown in Fig. \ref{fig:confusion_matrices}) and also a higher empirical risk for the segmentation network.

We also pre-train a diffusion model using a combination of unannotated images from all three datasets and then learn separate segmentation models for each dataset. 
From Table \ref{tab:combined_datasets}, we notice the segmentation performance on all three datasets is comparable to the case where both stages of our framework are trained on a single dataset.
In the case of HN cancer, we observe a slight improvement in one of the metrics (AJI) since the combined training data has about $70\%$ training samples coming from the HN cancer dataset.
In the case of GlaS, there is a slight improvement in performance in all metrics.
However, in the case of MoNuSeg, there is a slight improvement in only HD.
This validates that pre-training on a combination of datasets generalizes well without compromising the segmentation performance.

\begin{table}[!t]
\caption{Evaluation of Performance of Combining the Three Datasets GlaS, MoNuSeg, and HN Cancer During Self-Supervision.}
\begin{center}
\setlength{\tabcolsep}{3pt}
\begin{tabular}{p{90pt}p{30pt}p{40pt}p{40pt}}
\hline
Metric & GlaS & MoNuSeg & HN cancer\\
\hline
Aggregated Jaccard Index & 0.8543 & 0.6497 & 0.7931\\
Intersection over Union & 0.8543 & 0.6497 & 0.8034\\
Hausdorff Distance & 6.1941 & 7.3181 & 4.7540\\
F1-score & 0.9095 & 0.7852 & 0.8341\\
\hline
\end{tabular}
\label{tab:combined_datasets}
\end{center}
\end{table}

\begin{table}[!t]
\caption{Evaluation of Segmentation Performance using AJI on Cross-Dataset Self-Supervision. Row Indicates the Pre-trained Dataset.}
\begin{center}
\setlength{\tabcolsep}{3pt}
\begin{tabular}{p{45pt}p{30pt}p{40pt}p{40pt}}
\hline
Dataset & GlaS & MoNuSeg & HN cancer\\
\hline
GlaS & 0.8470 & 0.6514 & 0.7610\\
MoNuSeg & 0.8333 & \textbf{0.6545} & 0.7524\\
HN cancer & \textbf{0.8479} & 0.6432 & \textbf{0.7913}\\
\hline
\end{tabular}
\label{tab:cross_dataset}
\end{center}
\end{table}

\begin{table}[!t]
\caption{Effect of Number of Diffusion Time Steps During Pretraining on Segmentation Performance on GlaS and MoNuSeg Datasets.}
\begin{center}
\setlength{\tabcolsep}{3pt}
\begin{tabular}{p{45pt}p{30pt}p{40pt}}
\hline
Time stamps & GlaS & MoNuSeg \\
\hline
50 & 0.8431 & 0.6415 \\
100 & 0.8427 & 0.6486 \\
250 & 0.8428 & 0.6377 \\
500 & 0.8367 & 0.6248 \\
1000 & \textbf{0.8470} & \textbf{0.6545} \\
2000 & 0.8454 & 0.6381 \\
5000 & 0.8339 & 0.6338 \\
\hline
\end{tabular}
\label{tab:time_steps}
\end{center}
\end{table}

\subsection{Cross-Dataset Segmentation}
Table \ref{tab:datasets} shows that the HN cancer dataset contains the most unannotated images, followed by MoNuSeg and GLaS.
From Fig. \ref{fig:comparison_methods}, it can be seen that GLaS and HN cancer datasets have comparable mask sizes for any class in the annotated segmentation maps, whereas the mask size (corresponding to the nucleus class) is small. This indicates some amount of correlation between the two datasets.
Hence, from Table \ref{tab:cross_dataset}, we observe that the segmentation performance on the GLaS dataset using a model pre-trained on the HN dataset is good. 
However, self-supervision using the GLaS dataset and segmentation on the HN dataset does not follow the same pattern due to the low number of unannotated images in the GLaS dataset. 
The model achieves good performance on the MoNuSeg dataset when pre-trained and fine-tuned on MoNuSeg itself. This indicates that a certain degree of similarity between datasets aids in learning task-agnostic visual representations during self-supervision.

\section{Discussion}
\subsection{Ablations} 
\subsubsection{Effect of Diffusion Time Steps} We observe the effect of varying the number of time steps $T$ in the diffusion process during the pre-training stage on the performance of segmentation.
The time steps are varied from $50$ to $5000$.
Table \ref{tab:time_steps} shows the AJI values on MoNuSeg and GlaS datasets.
We notice that the performance degrades when $T$ is either too high or too low for both datasets.
When $T$ is high, the noise addition in the later steps is a very slow process and is less significant. 
As a result, most noising steps are high SNR processes, making the network learn many imperceptible details and content from these images.
Hence, a performance drop is expected.
On the other hand, when $T$ is very low, the noise addition happens very fast, adding significant noise in each step.
This causes the noising process to operate in a low SNR regime, making the model learn predominantly coarse details over content information, resulting in a performance drop \cite{b45}.
Hence, a trade-off exists based on the choice of $T$, impacting the segmentation performance.
We observe that the AJI scores roughly peak simultaneously at $T=1000$ for both datasets, which we use for all pre-training experiments.

\subsubsection{Effect of Loss Functions} \label{loss effects}
\begin{figure}[!t]
\centerline{\includegraphics[width=\columnwidth]{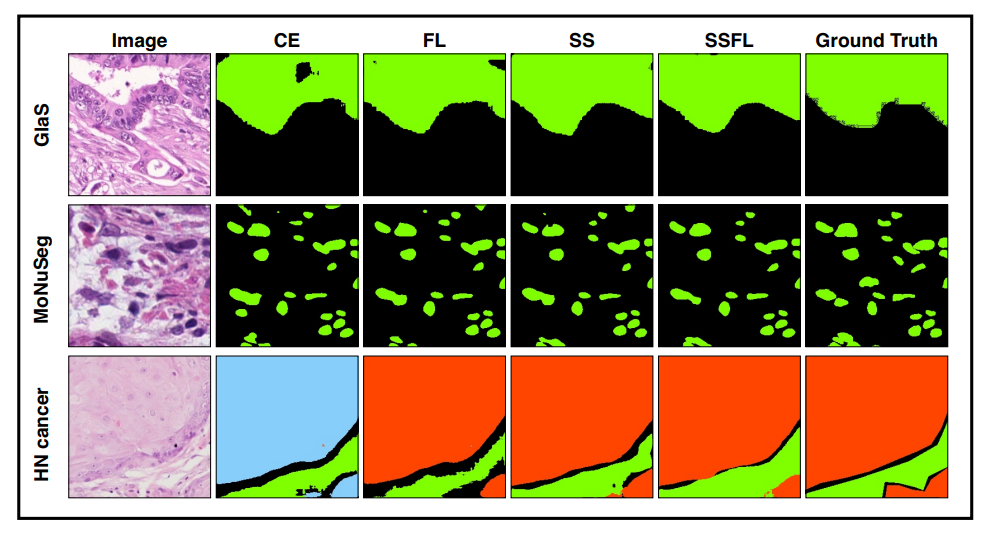}}
\caption{Qualitative Results of the proposed multi-loss function on all three datasets. The losses are abbreviated as CE - cross-entropy, FL - focal loss, and SS - structural similarity.}
\label{fig:comparison_multiloss}
\end{figure}

In our work, we use a combination of structural similarity and focal losses.
In Table \ref{tab:loss_functions}, we explore the contribution of individual losses: CE, SS, and FL.
We observe that CE loss performs well only on MoNuSeg, where the training data is less, and poorly on the other two datasets.
Qualitative results are presented in Fig. \ref{fig:comparison_multiloss} to show the effect of different loss functions on segmentation performance. When we have access to more training data, both SS and FL perform better individually.
Finally, the combination of SS and FL gives the best performance on all the segmentation datasets.

\subsubsection{Effect of Loss Scaling Factor} \label{subsec: ablation}
To evaluate the contribution of each loss term in our multi-loss function, we vary the hyperparameter $\lambda \in \{ 0.1, 0.5, 1, 10\}$.
From Table \ref{tab:ablat1}, we observe that an equal weighting of both the SS loss and focal loss ($\lambda = 1$) offers the best performance in terms of AJI, indicating that a balancing reconstruction error and class imbalance is necessary for efficient learning.

\begin{table}[t]
\caption{Effect of Multi Loss Function on All Three Datasets. CE, SS, and FL Indicate Cross Entropy, Structural Similarity, and Focal Losses, Respectively.}
\begin{center}
\setlength{\tabcolsep}{3pt}
\begin{tabular}{p{45pt}p{30pt}p{40pt}p{40pt}}
\hline
Loss & GlaS & MoNuSeg & HN cancer\\
\hline
CE & 0.7923 & 0.6529 & 0.7744\\
SS & 0.8093 & 0.6323 & 0.7636\\
FL & 0.8065 & 0.6510 & 0.7730\\
SS + FL & \textbf{0.8470} & \textbf{0.6545} & \textbf{0.7913}\\
\hline
\end{tabular}
\label{tab:loss_functions}
\end{center}
\end{table}

\begin{table}[t]
\caption{Evaluation of Segmentation Performance Using AJI on Varying Scale Factor of the Multi Loss Function.}
\begin{center}
\setlength{\tabcolsep}{3pt}
\begin{tabular}{p{45pt}p{30pt}p{40pt}p{40pt}}
\hline
Scale factor & GlaS & MoNuSeg & HN cancer\\
\hline
0.1 & 0.8206 & 0.6536 & 0.7634\\
0.5 & 0.8183 & 0.6445 & 0.7604\\
1.0 & \textbf{0.8470} & \textbf{0.6545} & \textbf{0.7913}\\
10 & 0.8154 & 0.6497 & 0.7754\\
\hline
\end{tabular}
\label{tab:ablat1}
\end{center}
\end{table}

\subsubsection{Effect of Varying Architecture Size}
We experiment with varying U-Net block sizes and record the effect of network size on the segmentation performance. 
The number of downsampling (or upsampling) levels varies between $3$ to $5$, along with the number of output channels in each level in the U-Net.
Table \ref{tab:ablat2} illustrates the number of levels accompanied by the number of output channels in each level.
It is observed that a U-Net consisting of $4$ levels with $(64, 128, 256, 512)$ output channels performs the best.
Moreover, within $4$ level U-Net architectures, we note that the standard UNet performs the best, indicating that this network efficiently trades off bias and overfitting.

\begin{table}[t]
\caption{Segmentation Performance Using AJI for Different Sizes of UNet Architecture. The Numbers in Brackets Indicate the Channels.}
\begin{center}
\setlength{\tabcolsep}{3pt}
\begin{tabular}{p{95pt}p{30pt}p{40pt}p{40pt}}
\hline
Blocks & GlaS & MoNuSeg\\
\hline
3 (64, 128, 256) & 0.8342 & 0.6396\\
4 (32, 64, 128, 256) & 0.8268 & 0.6442\\
4 (64, 128, 192, 256) & 0.8427 & 0.6468\\
4 (64, 128, 256, 512) & \textbf{0.8470} & \textbf{0.6545}\\
5 (64, 128, 256, 512, 1024) & 0.8187 & 0.6342\\
\hline
\end{tabular}
\label{tab:ablat2}
\end{center}
\end{table}

\subsection{Observations and Analysis}

Our two-stage framework allows us to learn efficient histopathological image representations via diffusion as a self-supervised pretext task. 
Since DDPMs inherently learn an image-to-image Markov process, they are the most suitable choice for learning good representations that are downstreamed for segmentation.
We observe from Table \ref{tab:generation_perf} that the generation performance of the DDPM is consistent with the number of unlabelled training examples.
In addition, this also leads to improved generalization (Table \ref{tab:cross_dataset}), where the performance in a cross-dataset setting is higher in the case where the DDPM is pre-trained on HN cancer and downstreamed on GlaS than the case where GlaS is used in both stages.
However, in the case where the datasets are combined for SSL pretraining, the performance slightly improves for GlaS in all metrics, whereas there is a slight drop in the cases of MoNuSeg and HN cancer in most metrics.
It is worth noting that a single SSL pre-trained model is downstreamed for all three datasets, and the segmentation performance is still comparable to pretraining on single datasets.
This validates that generalization is possible without having to compromise on the downstream performance.

From Table \ref{tab:main_table}, we see that our method outperforms other image-to-image pretext tasks as well as DDPM-based methods.
While our framework requires training a DDPM, it is worth noting that the $T$ time steps are only used during the pretraining phase, and not during inference. 
Since segmentation doesn't require the addition of noise or denoising of any kind, the DDPM at $t=0$ is used for downstream segmentation.
This significantly reduces the inference time compared to other methods like \cite{b74,b73,b72,b21} since these methods require the entire sampling process of the DDPM for predicting masks.
Moreover, since there is no noise involved after pretraining, our predicted segmentation masks neither have any ambiguity nor require an ensemble of predictions. 
As a result, our method (along with pretext tasks) performs better than the DDPM-based methods in almost all cases.

\subsection{GenSelfDiff for Object Detection}

\begin{table}[t]
\caption{Object Detection Evaluation on the Three Datasets in Terms of Mean Intersection over Union (mIoU).}
\begin{center}
\begin{tabular}{c|c|c|c}
\hline
Weight Initialization                               & MoNuSeg                       & GlaS                          & HN Cancer                     \\ \hline
Trained from scratch         & 0.5643 & 0.4551 & 0.5254 \\
Ours (diffusion pre-training) & \textbf{0.5820} & \textbf{0.4603} & \textbf{0.5334} \\ \hline
\end{tabular}
\label{tab:obj_detection}
\end{center}
\end{table}

Our initial experiments show that our method can be extended beyond segmentation, for tasks such as object detection.
We conduct experiments where we try to detect and localize nuclei present in histopathological images.
To obtain bounding box annotations for the nuclei, we use the ground-truth segmentation masks to extract object coordinates and locations.
We use two convolutional followed by fully connected layers on top of the UNet architecture to modify it for object detection. 
Two models are trained: one from scratch, and another initialized using weights from our self-supervised pre-training stage with the same loss function used in the popular YOLO-V1 model \cite{b87}. 
From Table \ref{tab:obj_detection}, we observe that our self-supervision improves the performance of downstream object detection in terms of mean intersection over union (mIoU) as well.
This demonstrates the generalizability of our method over multiple tasks.

\section{Conclusion}

In this work, we proposed a novel two-stage learning approach for segmenting histopathological images. 
The first stage consists of learning good visual representations of histopathological images through a generative diffusion-based SSL.
This is achieved through a DDPM trained with unannotated data.
Subsequently, the SSL pre-trained network is trained for segmentation on a target dataset.
Our approach achieves the best performance among all self-supervised pretext tasks and also outperforms other DDPM-based and fully supervised methods.
We also introduce a new head and neck cancer dataset consisting of annotated and unannotated histopathological images. 
The research community can use this dataset to develop machine learning algorithms to analyze and understand histopathological images for segmentation, detection, classification, and many other tasks.
Our future work includes enhancing self-supervised representation learning with segmentation-specific losses and using auxiliary tasks with supervised losses.
Our future work also involves extending our method for tasks beyond segmentation and exploring other self-supervised methods for effective analysis of clinical images.

\bibliographystyle{IEEEtran}
\bibliography{ref}

\begin{thebibliography}{10}
\providecommand{\url}[1]{#1}
\csname url@samestyle\endcsname
\providecommand{\newblock}{\relax}
\providecommand{\bibinfo}[2]{#2}
\providecommand{\BIBentrySTDinterwordspacing}{\spaceskip=0pt\relax}
\providecommand{\BIBentryALTinterwordstretchfactor}{4}
\providecommand{\BIBentryALTinterwordspacing}{\spaceskip=\fontdimen2\font plus
\BIBentryALTinterwordstretchfactor\fontdimen3\font minus \fontdimen4\font\relax}
\providecommand{\BIBforeignlanguage}[2]{{%
\expandafter\ifx\csname l@#1\endcsname\relax
\typeout{** WARNING: IEEEtran.bst: No hyphenation pattern has been}%
\typeout{** loaded for the language `#1'. Using the pattern for}%
\typeout{** the default language instead.}%
\else
\language=\csname l@#1\endcsname
\fi
#2}}
\providecommand{\BIBdecl}{\relax}
\BIBdecl

\bibitem{b56}
C.~Lu, M.~Mahmood, N.~Jha, and M.~Mandal, ``Automated segmentation of the melanocytes in skin histopathological images,'' \emph{IEEE Journal of Biomedical and Health Informatics}, vol.~17, no.~2, pp. 284--296, 2013.

\bibitem{b57}
------, ``A robust automatic nuclei segmentation technique for quantitative histopathological image analysis,'' \emph{Analytical and Quantitative Cytology and Histology}, vol.~34, pp. 296--308, 2012.

\bibitem{b58}
M.~Mete, X.~Xu, C.-Y. Fan, and G.~Shafirstein, ``Automatic delineation of malignancy in histopathological head and neck slides,'' in \emph{BMC Bioinformatics}, vol.~8, no.~7.\hskip 1em plus 0.5em minus 0.4em\relax BioMed Central, 2007, pp. 1--13.

\bibitem{b1}
M.~Springenberg, A.~Frommholz, M.~Wenzel, E.~Weicken, J.~Ma, and N.~Strodthoff, ``From modern cnns to vision transformers: Assessing the performance, robustness, and classification strategies of deep learning models in histopathology,'' \emph{Medical Image Analysis}, vol.~87, p. 102809, 2023.

\bibitem{b2}
O.~Ronneberger, P.~Fischer, and T.~Brox, ``U-net: Convolutional networks for biomedical image segmentation,'' in \emph{Medical Image Computing and Computer-Assisted Intervention}.\hskip 1em plus 0.5em minus 0.4em\relax Springer Intl. Publishing, 2015, pp. 234--241.

\bibitem{b59}
A.~Albayrak and G.~Bilgin, ``A hybrid method of superpixel segmentation algorithm and deep learning method in histopathological image segmentation,'' in \emph{Innovations in Intelligent Systems and Applications (INISTA)}.\hskip 1em plus 0.5em minus 0.4em\relax IEEE, 2018, pp. 1--5.

\bibitem{b60}
Y.~Xu, Z.~Jia, L.-B. Wang, Y.~Ai, F.~Zhang, M.~Lai, E.~I. Chang \emph{et~al.}, ``Large scale tissue histopathology image classification, segmentation, and visualization via deep convolutional activation features,'' \emph{BMC Bioinformatics}, vol.~18, no.~1, pp. 1--17, 2017.

\bibitem{b78}
A.~Tragakis, C.~Kaul, R.~Murray-Smith, and D.~Husmeier, ``The fully convolutional transformer for medical image segmentation,'' in \emph{Proceedings of the IEEE/CVF Winter Conference on Applications of Computer Vision}, 2023, pp. 3660--3669.

\bibitem{b61}
J.~Xu, ``A review of self-supervised learning methods in the field of medical image analysis,'' \emph{Intl. Journal of Image, Graphics and Signal Processing (IJIGSP)}, vol.~13, no.~4, pp. 33--46, 2021.

\bibitem{b62}
R.~Krishnan, P.~Rajpurkar, and E.~J. Topol, ``Self-supervised learning in medicine and healthcare,'' \emph{Nature Biomedical Engineering}, vol.~6, no.~12, pp. 1346--1352, 2022.

\bibitem{b3}
R.~Zhang, P.~Isola, and A.~A. Efros, ``Split-brain autoencoders: Unsupervised learning by cross-channel prediction,'' in \emph{IEEE Conf. on Computer Vision and Pattern Recognition (CVPR)}, 2017, pp. 645--654.

\bibitem{b4}
L.~Chen, P.~Bentley, K.~Mori, K.~Misawa, M.~Fujiwara, and D.~Rueckert, ``Self-supervised learning for medical image analysis using image context restoration,'' \emph{Medical Image Analysis}, vol.~58, p. 101539, 2019.

\bibitem{b5}
S.~Gidaris, P.~Singh, and N.~Komodakis, ``Unsupervised representation learning by predicting image rotations,'' in \emph{Intl. Conf. on Learning Representations}, 2018.

\bibitem{b6}
R.~Zhang, P.~Isola, and A.~A. Efros, ``Colorful image colorization,'' in \emph{Computer Vision -- ECCV 2016}.\hskip 1em plus 0.5em minus 0.4em\relax Springer Intl. Publishing, 2016, pp. 649--666.

\bibitem{b7}
C.~Ledig, L.~Theis, F.~Huszár, J.~Caballero, A.~Cunningham, A.~Acosta, A.~Aitken, A.~Tejani, J.~Totz, Z.~Wang, and W.~Shi, ``Photo-realistic single image super-resolution using a generative adversarial network,'' in \emph{IEEE Conf. on Computer Vision and Pattern Recognition (CVPR)}, 2017, pp. 105--114.

\bibitem{b8}
D.~Pathak, P.~Krahenbuhl, J.~Donahue, T.~Darrell, and A.~A. Efros, ``Context encoders: Feature learning by inpainting,'' in \emph{IEEE Conf. on Computer Vision and Pattern Recognition (CVPR)}, jun 2016, pp. 2536--2544.

\bibitem{b9}
C.~L. Srinidhi, S.~W. Kim, F.-D. Chen, and A.~L. Martel, ``Self-supervised driven consistency training for annotation efficient histopathology image analysis,'' \emph{Medical Image Analysis}, vol.~75, p. 102256, 2022.

\bibitem{b10}
P.~Chhipa, R.~Upadhyay, G.~Pihlgren, R.~Saini, S.~Uchida, and M.~Liwicki, ``Magnification prior: A self-supervised method for learning representations on breast cancer histopathological images,'' in \emph{IEEE/CVF Winter Conf. on Applications of Computer Vision (WACV)}.\hskip 1em plus 0.5em minus 0.4em\relax IEEE Computer Society, jan 2023, pp. 2716--2726.

\bibitem{b11}
J.~Xu, J.~Hou, Y.~Zhang, R.~Feng, C.~Ruan, T.~Zhang, and W.~Fan, ``Data-efficient histopathology image analysis with deformation representation learning,'' in \emph{IEEE Intl. Conf. on Bioinformatics and Biomedicine (BIBM)}, 2020, pp. 857--864.

\bibitem{b12}
X.~Wang, S.~Yang, J.~Zhang, M.~Wang, J.~Zhang, J.~Huang, W.~Yang, and X.~Han, ``Transpath: Transformer-based self-supervised learning for histopathological image classification,'' in \emph{Medical Image Computing and Computer Assisted Intervention}.\hskip 1em plus 0.5em minus 0.4em\relax Springer Intl. Publishing, 2021, pp. 186--195.

\bibitem{b13}
C.~Abbet, I.~Zlobec, B.~Bozorgtabar, and J.-P. Thiran, ``Divide-and-rule: Self-supervised learning for survival analysis in colorectal cancer,'' in \emph{Medical Image Computing and Computer Assisted Intervention}.\hskip 1em plus 0.5em minus 0.4em\relax Springer Intl. Publishing, 2020, pp. 480--489.

\bibitem{b41}
P.~Yang, X.~Yin, H.~Lu, Z.~Hu, X.~Zhang, R.~Jiang, and H.~Lv, ``Cs-co: A hybrid self-supervised visual representation learning method for h\&e-stained histopathological images,'' \emph{Medical Image Analysis}, vol.~81, p. 102539, 2022.

\bibitem{b14}
D.~Mahapatra, B.~Bozorgtabar, J.-P. Thiran, and L.~Shao, ``Structure preserving stain normalization of histopathology images using self supervised semantic guidance,'' in \emph{Medical Image Computing and Computer Assisted Intervention}.\hskip 1em plus 0.5em minus 0.4em\relax Springer Intl. Publishing, 2020, pp. 309--319.

\bibitem{b15}
A.~C. Quiros, R.~Murray-Smith, and K.~Yuan, ``Pathologygan: Learning deep representations of cancer tissue,'' \emph{Machine Learning for Biomedical Imaging}, vol.~1, pp. 1--47, 2021.

\bibitem{b16}
A.~Claudio~Quiros, N.~Coudray, A.~Yeaton, W.~Sunhem, R.~Murray-Smith, A.~Tsirigos, and K.~Yuan, ``Adversarial learning of cancer tissue representations,'' in \emph{Medical Image Computing and Computer Assisted Intervention}.\hskip 1em plus 0.5em minus 0.4em\relax Springer Intl. Publishing, 2021, pp. 602--612.

\bibitem{b17}
Y.~Zhao, F.~Yang, Y.~Fang, H.~Liu, N.~Zhou, J.~Zhang, J.~Sun, S.~Yang, B.~Menze, X.~Fan, and J.~Yao, ``Predicting lymph node metastasis using histopathological images based on multiple instance learning with deep graph convolution,'' in \emph{IEEE/CVF Conf. on Computer Vision and Pattern Recognition (CVPR)}, 2020, pp. 4836--4845.

\bibitem{b18}
D.~P. Kingma and M.~Welling, ``{Auto-Encoding Variational Bayes},'' in \emph{Intl. Conf. on Learning Representations}, 2014.

\bibitem{b19}
I.~Goodfellow, J.~Pouget-Abadie, M.~Mirza, B.~Xu, D.~Warde-Farley, S.~Ozair, A.~Courville, and Y.~Bengio, ``Generative adversarial nets,'' in \emph{Advances in Neural Information Processing Systems}, vol.~27.\hskip 1em plus 0.5em minus 0.4em\relax Curran Associates, Inc., 2014.

\bibitem{b63}
K.~Zhang, ``On mode collapse in generative adversarial networks,'' in \emph{Artificial Neural Networks and Machine Learning--ICANN 2021: 30th Intl. Conf. on Artificial Neural Networks}.\hskip 1em plus 0.5em minus 0.4em\relax Springer, 2021, pp. 563--574.

\bibitem{b44}
J.~Ho, A.~Jain, and P.~Abbeel, ``Denoising diffusion probabilistic models,'' \emph{Advances in Neural Information Processing Systems}, vol.~33, pp. 6840--6851, 2020.

\bibitem{b20}
P.~Dhariwal and A.~Q. Nichol, ``Diffusion models beat {GAN}s on image synthesis,'' in \emph{Advances in Neural Information Processing Systems}, 2021.

\bibitem{b21}
J.~Wu, R.~FU, H.~Fang, Y.~Zhang, Y.~Yang, H.~Xiong, H.~Liu, and Y.~Xu, ``Medsegdiff: Medical image segmentation with diffusion probabilistic model,'' in \emph{Medical Imaging with Deep Learning}, 2023.

\bibitem{b22}
J.~Wu, R.~Fu, H.~Fang, Y.~Zhang, and Y.~Xu, ``Medsegdiff-v2: Diffusion based medical image segmentation with transformer,'' \emph{ArXiv}, vol. abs/2301.11798, 2023.

\bibitem{b23}
C.~L. Srinidhi, O.~Ciga, and A.~L. Martel, ``Deep neural network models for computational histopathology: A survey,'' \emph{Medical Image Analysis}, vol.~67, p. 101813, 2021.

\bibitem{b24}
D.~Komura and S.~Ishikawa, ``Machine learning methods for histopathological image analysis,'' \emph{Computational and Structural Biotechnology Journal}, vol.~16, pp. 34--42, 2018.

\bibitem{b25}
Z.~Zhou, M.~M.~R. Siddiquee, N.~Tajbakhsh, and J.~Liang, ``Unet++: Redesigning skip connections to exploit multiscale features in image segmentation,'' \emph{IEEE Transactions on Medical Imaging}, vol.~39, no.~6, pp. 1856--1867, 2020.

\bibitem{b26}
K.~He, X.~Zhang, S.~Ren, and J.~Sun, ``Deep residual learning for image recognition,'' in \emph{IEEE Conf. on Computer Vision and Pattern Recognition (CVPR)}, 2016, pp. 770--778.

\bibitem{b27}
Y.~Xu, J.-Y. Zhu, E.~I.-C. Chang, M.~Lai, and Z.~Tu, ``Weakly supervised histopathology cancer image segmentation and classification,'' \emph{Medical Image Analysis}, vol.~18, no.~3, pp. 591--604, 2014.

\bibitem{b28}
H.~Chen, X.~Qi, L.~Yu, and P.-A. Heng, ``Dcan: Deep contour-aware networks for accurate gland segmentation,'' in \emph{IEEE Conf. on Computer Vision and Pattern Recognition (CVPR)}, 2016, pp. 2487--2496.

\bibitem{b29}
Y.~Liu, Q.~He, H.~Duan, H.~Shi, A.~Han, and Y.~He, ``Using sparse patch annotation for tumor segmentation in histopathological images,'' \emph{Sensors}, vol.~22, no.~16, 2022.

\bibitem{b30}
J.~Bokhorst, H.~Pinckaers, P.~{van Zwam}, I.~Nagtegaal, J.~{van der Laak}, and F.~Ciompi, ``Learning from sparsely annotated data for semantic segmentation in histopathology images,'' in \emph{Proc. of The 2nd Intl. Conf. on Medical Imaging with Deep Learning}, vol. 102.\hskip 1em plus 0.5em minus 0.4em\relax PMLR, 2019, pp. 84--91.

\bibitem{b31}
J.~Yan, H.~Chen, X.~Li, and J.~Yao, ``Deep contrastive learning based tissue clustering for annotation-free histopathology image analysis,'' \emph{Computerized Medical Imaging and Graphics}, vol.~97, p. 102053, 2022.

\bibitem{b32}
P.~Yang, Y.~Zhai, L.~Li, H.~Lv, J.~Wang, C.~Zhu, and R.~Jiang, ``A deep metric learning approach for histopathological image retrieval,'' \emph{Methods}, vol. 179, pp. 14--25, 2020, interpretable Machine Learning in Bioinformatics.

\bibitem{b33}
M.~N. Gurcan, L.~E. Boucheron, A.~Can, A.~Madabhushi, N.~M. Rajpoot, and B.~Yener, ``Histopathological image analysis: A review,'' \emph{IEEE Reviews in Biomedical Engineering}, vol.~2, pp. 147--171, 2009.

\bibitem{b34}
L.~Qu, S.~Liu, X.~Liu, M.~Wang, and Z.~Song, ``Towards label-efficient automatic diagnosis and analysis: a comprehensive survey of advanced deep learning-based weakly-supervised, semi-supervised and self-supervised techniques in histopathological image analysis,'' \emph{Physics in Medicine \& Biology}, vol.~67, no.~20, p. 20TR01, oct 2022.

\bibitem{b35}
L.~Jing and Y.~Tian, ``Self-supervised visual feature learning with deep neural networks: A survey,'' \emph{IEEE Transactions on Pattern Analysis and Machine Intelligence}, vol.~43, no.~11, pp. 4037--4058, 2021.

\bibitem{b36}
X.~Liu, F.~Zhang, Z.~Hou, L.~Mian, Z.~Wang, J.~Zhang, and J.~Tang, ``Self-supervised learning: Generative or contrastive,'' \emph{IEEE Transactions on Knowledge and Data Engineering}, vol.~35, no.~1, pp. 857--876, 2023.

\bibitem{b37}
N.~A. Koohbanani, B.~Unnikrishnan, S.~A. Khurram, P.~Krishnaswamy, and N.~Rajpoot, ``Self-path: Self-supervision for classification of pathology images with limited annotations,'' \emph{IEEE Transactions on Medical Imaging}, vol.~40, no.~10, pp. 2845--2856, 2021.

\bibitem{b39}
T.~Chen, S.~Kornblith, M.~Norouzi, and G.~Hinton, ``A simple framework for contrastive learning of visual representations,'' in \emph{Proceedings of the 37th Intl. Conf. on Machine Learning}, ser. Proceedings of Machine Learning Research, vol. 119.\hskip 1em plus 0.5em minus 0.4em\relax PMLR, 13--18 Jul 2020, pp. 1597--1607.

\bibitem{b54}
K.~He, H.~Fan, Y.~Wu, S.~Xie, and R.~Girshick, ``Momentum contrast for unsupervised visual representation learning,'' in \emph{Proceedings of the IEEE/CVF Conf. on computer vision and pattern recognition}, 2020, pp. 9729--9738.

\bibitem{b55}
J.-B. Grill, F.~Strub, F.~Altch{\'e}, C.~Tallec, P.~Richemond, E.~Buchatskaya, C.~Doersch, B.~Avila~Pires, Z.~Guo, M.~Gheshlaghi~Azar \emph{et~al.}, ``Bootstrap your own latent-a new approach to self-supervised learning,'' \emph{Advances in neural information processing systems}, vol.~33, pp. 21\,271--21\,284, 2020.

\bibitem{b38}
K.~Chaitanya, E.~Erdil, N.~Karani, and E.~Konukoglu, ``Contrastive learning of global and local features for medical image segmentation with limited annotations,'' in \emph{Advances in Neural Information Processing Systems}, vol.~33.\hskip 1em plus 0.5em minus 0.4em\relax Curran Associates, Inc., 2020, pp. 12\,546--12\,558.

\bibitem{b40}
O.~Ciga, T.~Xu, and A.~L. Martel, ``Self supervised contrastive learning for digital histopathology,'' \emph{Machine Learning with Applications}, vol.~7, p. 100198, 2022.

\bibitem{b42}
K.~Stacke, J.~Unger, C.~Lundström, and G.~Eilertsen, ``Learning representations with contrastive self-supervised learning for histopathology applications,'' \emph{Machine Learning for Biomedical Imaging}, vol.~1, pp. 1--33, 2022.

\bibitem{b71}
T.~Amit, T.~Shaharbany, E.~Nachmani, and L.~Wolf, ``Segdiff: Image segmentation with diffusion probabilistic models,'' \emph{arXiv preprint arXiv:2112.00390}, 2021.

\bibitem{b74}
D.~Baranchuk, A.~Voynov, I.~Rubachev, V.~Khrulkov, and A.~Babenko, ``Label-efficient semantic segmentation with diffusion models,'' in \emph{International Conference on Learning Representations}, 2022.

\bibitem{b73}
J.~Wolleb, R.~Sandk{\"u}hler, F.~Bieder, P.~Valmaggia, and P.~C. Cattin, ``Diffusion models for implicit image segmentation ensembles,'' in \emph{International Conference on Medical Imaging with Deep Learning}.\hskip 1em plus 0.5em minus 0.4em\relax PMLR, 2022, pp. 1336--1348.

\bibitem{b75}
B.~Kim, Y.~Oh, and J.~C. Ye, ``Diffusion adversarial representation learning for self-supervised vessel segmentation,'' in \emph{The Eleventh International Conference on Learning Representations}, 2023.

\bibitem{b72}
A.~Rahman, J.~M.~J. Valanarasu, I.~Hacihaliloglu, and V.~M. Patel, ``Ambiguous medical image segmentation using diffusion models,'' in \emph{Proceedings of the IEEE/CVF Conference on Computer Vision and Pattern Recognition}, 2023, pp. 11\,536--11\,546.

\bibitem{b76}
F.-A. Croitoru, V.~Hondru, R.~T. Ionescu, and M.~Shah, ``Diffusion models in vision: A survey,'' \emph{IEEE Transactions on Pattern Analysis and Machine Intelligence}, 2023.

\bibitem{b77}
A.~Kazerouni, E.~K. Aghdam, M.~Heidari, R.~Azad, M.~Fayyaz, I.~Hacihaliloglu, and D.~Merhof, ``Diffusion models in medical imaging: A comprehensive survey,'' \emph{Medical Image Analysis}, p. 102846, 2023.

\bibitem{b45}
J.~Choi, J.~Lee, C.~Shin, S.~Kim, H.~Kim, and S.~Yoon, ``Perception prioritized training of diffusion models,'' in \emph{IEEE/CVF Conf. on Computer Vision and Pattern Recognition (CVPR)}, 2022, pp. 11\,462--11\,471.

\bibitem{b43}
C.~Luo, ``Understanding diffusion models: A unified perspective,'' \emph{ArXiv}, vol. abs/2208.11970, 2022.

\bibitem{b48}
S.~Zhao, B.~Wu, W.~Chu, Y.~Hu, and D.~Cai, ``Correlation maximized structural similarity loss for semantic segmentation,'' \emph{ArXiv}, vol. abs/1910.08711, 2019.

\bibitem{b68}
T.-Y. Lin, P.~Goyal, R.~Girshick, K.~He, and P.~Doll{\'a}r, ``Focal loss for dense object detection,'' in \emph{Proceedings of the IEEE Intl. Conf. on Computer Vision}, 2017, pp. 2980--2988.

\bibitem{b49}
S.~Jadon, ``A survey of loss functions for semantic segmentation,'' in \emph{IEEE Conf. on Computational Intelligence in Bioinformatics and Computational Biology (CIBCB)}, 2020, pp. 1--7.

\bibitem{b51}
K.~Sirinukunwattana, J.~P. Pluim, H.~Chen, X.~Qi, P.-A. Heng, Y.~B. Guo, L.~Y. Wang, B.~J. Matuszewski, E.~Bruni, U.~Sanchez \emph{et~al.}, ``Gland segmentation in colon histology images: The glas challenge contest,'' \emph{Medical Image Analysis}, vol.~35, pp. 489--502, 2017.

\bibitem{b50}
N.~Kumar, R.~Verma, S.~Sharma, S.~Bhargava, A.~Vahadane, and A.~Sethi, ``A dataset and a technique for generalized nuclear segmentation for computational pathology,'' \emph{IEEE Transactions on Medical Imaging}, vol.~36, no.~7, pp. 1550--1560, 2017.

\bibitem{b65}
A.~J. Kimple, C.~M. Welch, J.~P. Zevallos, and S.~N. Patel, ``Oral cavity squamous cell carcinoma—an overview,'' \emph{Oral Health Dent Manag}, vol.~13, no.~3, pp. 877--882, 2014.

\bibitem{b79}
A.~Srivastava, S.~Sengupta, S.-J. Kang, K.~Kant, M.~Khan, S.~A. Ali, S.~R. Moore, B.~C. Amadi, P.~Kelly, S.~Syed \emph{et~al.}, ``Deep learning for detecting diseases in gastrointestinal biopsy images,'' in \emph{2019 Systems and information engineering design symposium (SIEDS)}.\hskip 1em plus 0.5em minus 0.4em\relax IEEE, 2019, pp. 1--4.

\bibitem{b80}
D.~Komura and S.~Ishikawa, ``Machine learning methods for histopathological image analysis,'' \emph{Computational and structural biotechnology journal}, vol.~16, pp. 34--42, 2018.

\bibitem{b81}
M.~Halicek, M.~Shahedi, J.~V. Little, A.~Y. Chen, L.~L. Myers, B.~D. Sumer, and B.~Fei, ``Head and neck cancer detection in digitized whole-slide histology using convolutional neural networks,'' \emph{Scientific reports}, vol.~9, no.~1, p. 14043, 2019.

\bibitem{b82}
W.~Bulten, H.~Pinckaers, H.~van Boven, R.~Vink, T.~de~Bel, B.~van Ginneken, J.~van~der Laak, C.~Hulsbergen-van~de Kaa, and G.~Litjens, ``Automated deep-learning system for gleason grading of prostate cancer using biopsies: a diagnostic study,'' \emph{The Lancet Oncology}, vol.~21, no.~2, pp. 233--241, 2020.

\bibitem{b83}
T.~Y. Rahman, L.~B. Mahanta, A.~K. Das, and J.~D. Sarma, ``Histopathological imaging database for oral cancer analysis,'' \emph{Data in brief}, vol.~29, p. 105114, 2020.

\bibitem{b84}
M.~M.~R. Krishnan, V.~Venkatraghavan, U.~R. Acharya, M.~Pal, R.~R. Paul, L.~C. Min, A.~K. Ray, J.~Chatterjee, and C.~Chakraborty, ``Automated oral cancer identification using histopathological images: a hybrid feature extraction paradigm,'' \emph{Micron}, vol.~43, no. 2-3, pp. 352--364, 2012.

\bibitem{b85}
S.~Sharma and R.~Mehra, ``Conventional machine learning and deep learning approach for multi-classification of breast cancer histopathology images—a comparative insight,'' \emph{Journal of digital imaging}, vol.~33, no.~3, pp. 632--654, 2020.

\bibitem{b86}
F.~A. Spanhol, L.~S. Oliveira, P.~R. Cavalin, C.~Petitjean, and L.~Heutte, ``Deep features for breast cancer histopathological image classification,'' in \emph{2017 IEEE International Conference on Systems, Man, and Cybernetics (SMC)}.\hskip 1em plus 0.5em minus 0.4em\relax IEEE, 2017, pp. 1868--1873.

\bibitem{b66}
O.~Oktay, J.~Schlemper, L.~Le~Folgoc, M.~Lee, M.~Heinrich, K.~Misawa, K.~Mori, S.~McDonagh, N.~Y. Hammerla, B.~Kainz \emph{et~al.}, ``Attention u-net: Learning where to look for the pancreas,'' in \emph{Medical Imaging with Deep Learning}, 2022.

\bibitem{b52}
R.~D. Hjelm, A.~Fedorov, S.~Lavoie-Marchildon, K.~Grewal, P.~Bachman, A.~Trischler, and Y.~Bengio, ``Learning deep representations by mutual information estimation and maximization,'' in \emph{Intl. Conf. on Learning Representations}, 2018.

\bibitem{b69}
J.-Y. Zhu, T.~Park, P.~Isola, and A.~A. Efros, ``Unpaired image-to-image translation using cycle-consistent adversarial networks,'' in \emph{Proceedings of the IEEE international conference on computer vision}, 2017, pp. 2223--2232.

\bibitem{b46}
A.~Q. Nichol and P.~Dhariwal, ``Improved denoising diffusion probabilistic models,'' in \emph{Proceedings of the 38th Intl. Conf. on Machine Learning}, ser. Proceedings of Machine Learning Research, vol. 139.\hskip 1em plus 0.5em minus 0.4em\relax PMLR, 18--24 Jul 2021, pp. 8162--8171.

\bibitem{b47}
P.~A. Moghadam, S.~Van~Dalen, K.~C. Martin, J.~Lennerz, S.~Yip, H.~Farahani, and A.~Bashashati, ``A morphology focused diffusion probabilistic model for synthesis of histopathology images,'' in \emph{IEEE/CVF Winter Conf. on Applications of Computer Vision (WACV)}, 2023, pp. 1999--2008.

\bibitem{b67}
M.~Heusel, H.~Ramsauer, T.~Unterthiner, B.~Nessler, and S.~Hochreiter, ``Gans trained by a two time-scale update rule converge to a local nash equilibrium,'' \emph{Advances in Neural Information Processing Systems}, vol.~30, 2017.

\bibitem{b87}
J.~Redmon, S.~Divvala, R.~Girshick, and A.~Farhadi, ``You only look once: Unified, real-time object detection,'' in \emph{Proceedings of the IEEE conference on computer vision and pattern recognition}, 2016, pp. 779--788.

\end{thebibliography}

\end{document}